\documentclass[twoside,11pt]{article}
\usepackage{jmlr2e}

\usepackage[utf8]{inputenc}
\usepackage{graphicx}
\usepackage{multirow}
\usepackage{hhline}
\usepackage{enumitem}
\usepackage{amssymb,amsmath,amsfonts,enumerate,float}
\usepackage{graphicx}
\usepackage{subcaption}
\usepackage[ruled,norelsize,linesnumbered]{algorithm2e}
\usepackage{bbm}
\usepackage{xcolor}
\usepackage{times}
\usepackage{booktabs}
\usepackage{hhline}
\usepackage{enumitem}

\usepackage[english]{babel}

\newcommand{\removelatexerror}{\let\@latex@error\@gobble}
\usepackage{bbm}

\newtheorem{thm}{Theorem}
\newtheorem{defn}{Definition}

\jmlrheading{1}{2000}{0--1}{0/00}{0/00}{jin00}{Yuan Jin, Wray Buntine, Fran{\c{c}}ois Petitjean and Geoffrey I.\ Webb}

\begin{document}

\title{Discriminative, Generative and Self-Supervised Approaches for Target-Agnostic Learning}
\author{Yuan Jin \and Wray Buntine \and Fran{\c{c}}ois Petitjean\thanks{Now at Australian Taxation Office.} \and Geoffrey I.\ Webb \\
\addr{Department of AI and Data Science, Faculty of IT} \\
\addr{Monash University, Clayton, Victoria, Australia}}

\editor{Unknown}

\maketitle

\begin{abstract}
Supervised learning, characterized by both discriminative and generative learning, seeks to predict the values of single (or sometimes multiple) predefined target attributes based on a predefined set of predictor attributes.
For applications where the information available and predictions to be made may vary from instance to instance, we propose the task of \textit{target-agnostic learning} where arbitrary disjoint sets of attributes can be used for each of predictors and targets for each to-be-predicted instance. For this task, we survey a wide range of techniques available for handling missing values, self-supervised training and pseudo-likelihood training, and adapt them to a suite of algorithms that are suitable for the task. We conduct extensive experiments on this suite of algorithms on a large collection of categorical, continuous and discretized datasets, and report their performance in terms of both classification and regression errors. We also report the training and prediction time of these algorithms when handling large-scale datasets.
Both generative and self-supervised learning models are shown to perform well at the task,
although their characteristics towards the different types of data are quite different. Nevertheless, our derived theorem for the pseudo-likelihood theory also shows that they are related for inferring a joint distribution model based on the pseudo-likelihood training.
\end{abstract}

\section{Introduction}

In many real-world situations, including medical diagnosis, customer relationship management, fraud prediction and sensor networks, both what we know and what we wish to infer can change dramatically depending on the context. For example, consider an autonomous AI system that is confronted by a combination of sensor failures that was never anticipated when the system was designed. Such a system would greatly benefit from a capacity to infer the information that is required for continuing to act from the sensors that are still operating; that is to use a learned model to predict values for attributes that were never expected to be targets from attributes that may not have been anticipated to be used as inputs. We do not want a self-driving car confronting an unexpected form of accident to just shut down because it does not have the information required to keep operating. Other more common cases include when a doctor wants to know the most likely results for different sets of measurements, tests and medications to be taken by different patients in advance, or when a bank advisor wants to predict the likelihood of different clients to be interested in different sets of products, based on the information collected for each  individual. 

Most approaches to supervised learning for tabular data create models that take a pre-specified set of attributes as inputs and a single pre-specified attribute as the target. This paper surveys a range of approaches to machine learning that create models that allow the input and target attributes to be specified at prediction time. We call this \emph{target-agnostic learning}.

One scenario in which a form of target agnostic learning has been studied is missing value imputation. This is a special case where the specification of the input and target attributes is determined by the vagaries of the data collection process. Target-agnostic learning creates unified models that are able to impute missing values at both training and prediction time, which has previously been performed separately~\citep{saar2007handling}. The unified models are characterized by the capacity of target-agnostic learning in marginalizing the joint distribution over any attributes, which will be elaborated in the following sections. Furthermore, it seems credible that other applications of target-agnostic learning can be very different from typical missing value imputation scenarios in terms of how the input and target attributes are specified.

Another scenario in which target-agnostic learning has received some attention is for generative graphical model learning~\citep{PetitjeanWebbTut16}. Generative graphical models, such as Bayesian network classifiers (BNCs)~\citep{webb2012learning}, hidden Markov models (HMMs)~\citep{harshvardhan2020comprehensive} and undirected graphical models~\citep{petitjean2013scaling,PetitjeanEtAl14a}, aim to learn the joint distribution of attributes by modelling the inter-dependencies between them. They are able to draw predictions for any target attributes conditioned on available evidence over the observed predictor attributes through marginalization of any other unobserved ones. The marginalization is enabled by exact or approximate inference algorithms. 
However, the focus in this area of research has been on accurate modelling of the full joint probability distribution.
The case of discriminative graphical models has shown that traditional graphical models, tuned for modelling the full joint distribution, are less accurate at predicting individual attributes than systems that are carefully regularized with respect to a target attribute.
These approaches, such as conditional random fields (CRFs)~\citep{sutton2006introduction}, gain considerable advantage relative to the generative approaches on making predictions for a predefined target. 
Some perform regularisation tailored to a single predefined target on its conditional log-likelihood~\citep{zaidi2013alleviating,zaidi2017efficient}. Others alter the graph structure using discriminative objectives to maximise the relationships between the (selected) set of predictors and the target~\citep{martinez2016scalable}. This suggests that target-agnostic graphical models that adjust inference depending on the target attributes are likely to have greater accuracy on those targets than current generative approaches.

More recently, self supervised learning in natural language processing
\citep{qiu2020pretrained} provides a further scenario for target-agnostic learning.  The
training for transformer language models such as BERT \citep{devlin2018bert} and XLNet \citep{yang2019xlnet} 
can be viewed as a form of target-agnostic learning because words in documents are individually 
predicted conditioned on other words during the training.  However, text, as well as images, appears to have homogeneous attribute types that are different from other domains which can have a complex mixture of attribute types (e.g. categorical, continuous or discretized attributes). For these domains, target-agnostic learning approaches can still be applied for effective prediction in the form of, for example, denoising autoencoders (DAEs) \citep{vincent2008extracting} and generative adversarial networks (GANs) \citep{Goodfellow2014gan}.

Recommender systems \citep{koren2015advances} can be viewed as another specialised subcase of target-agnostic learning. They focus on the scenario where the data is largely homogeneous (e.g. as integer ratings) and sparse as most users have not previously encountered most items.

The above are specialized cases of target-agnostic learning which have been summarized in Table \ref{tab:target_agnostic_summary}. In comparison, multi-task, multi-class and multi-label learning are related but different scenarios. They deal with learning multiple related tasks (possibly with different types of outputs), which ``transfers'' information among the tasks by means such as regularisation or parameter sharing. For target-agnostic learning, the multiple target attributes are neither required nor predetermined as in those three types of learning scenarios.


\begin{table*}[t]
\centering
\scriptsize
\begin{tabular}{|c|l|l|l|l|c|} 
\hline
 &\multicolumn{1}{|c|}{\textbf{Characteristics}}&\multicolumn{3}{|c|}{\textbf{Related Machine Learning Approaches}}&\textbf{Specialised}\\
 &&Generative&Discriminative&Self-Supervised&\textbf{Subcases}\\
 \hhline{~-----}
 \textbf{Target-Agnostic}&1. Arbitrary set of targets&1. BNCs&1. Discriminative&1. BERT&1. Missing value\\
 \textbf{Learning}&2. Arbitrary set of predictors&2. HMMs&~~~~BNCs&2. XLNet&~~~~imputation\\
 &3. Can be discriminative, &3. Junction tree&2. CRFs&3. DAEs&2. Recommender\\
 &~~~~generative or self-supervised&~~~~~models&&4. GANs&~~~~~systems\\
\hline
\end{tabular}
\caption{A summary of the characteristics of target-agnostic learning, its specialised subcases and related machine learning approaches.}\label{tab:target_agnostic_summary}
\end{table*}

Overall, due to the limitations of  existing research work, we believe that there is a need for further study of target-agnostic learning. Therefore, in this survey, we consider state-of-the-art approaches in graphical modelling, missing value imputation and self-supervised learning, and adapt them to target-agnostic learning. 

The survey is structured as follows. In Section 2, we review three machine learning paradigms connected with target-agnostic learning which are discriminative, generative and self-supervised learning. We then review the pseudo-likelihood theory which serves as the foundation of self-supervised learning approaches. For this theory, we also derive a theorem which shows that the above three learning paradigms are related with respect to inferring a joint distribution model based on the pseudo-likelihood training. 

In Section 3, we present a formal definition of target-agnostic learning, followed by the introduction of its specialised subcase, i.e. missing value imputation, and approaches that are either based on generative graphical models or adapted from self-supervised learning models with pseudo-likelihood training. 

In Section 4, we conduct a series of experiments that compare the different target-agnostic learning approaches across three types of data, which are the categorical, continuous and discretized data, with respect to classification and regression errors. We then report the training and prediction time of each approach in two scenarios: on large-quantity data and on a large number of attributes. We discuss in detail the performance of the different approaches and present our conjectures on why certain approaches perform better (or worse) in certain scenarios. 

In Section 5, we draw the conclusion with a summary of all the findings we have obtained from the experiments and the contributions we have made throughout the paper. Finally, we specify the future research directions based on this work.

\section{Related Work}

Target-agnostic learning has previously received very limited attention~\citep{capdevila2018experiments} and is not yet recognised as a distinct machine learning paradigm that requires specialised methodologies. It cannot be conveniently addressed by techniques from any existing paradigm of machine learning. Nevertheless, it is still important to articulate the connections between target-agnostic learning and the learning paradigms relevant to it in order for us to provide a formal definition and specialised approaches to it.

\subsection{Discriminative Learning}

Most approaches to supervised learning can be categorized into two paradigms: discriminative and generative learning. Both focus on drawing predictions about a
single pre-defined target attribute. Given a data instance described by a fixed set of attributes $\boldsymbol{V}$, discriminative models predict the target attribute $Y \in \boldsymbol{V}$ by estimating its conditional probability $P_{\boldsymbol{\theta}}(Y|\boldsymbol{X})$, where $\boldsymbol{X}=\boldsymbol{V} \setminus \{Y\}$ is the set of input attributes and $\boldsymbol{\theta}$ is the set of parameters of the conditional probability. This is also equivalent to learning a decision function $f_{\boldsymbol{\theta}}(Y;\boldsymbol{X})$ that maps the inputs $\boldsymbol{X}$ onto the target $Y$.  

To address the target-agnostic prediction task, discriminative learning needs to be able to predict, for the $i$-th new data instance, its subset of attributes $\boldsymbol{Y}^{(i)} \subset \boldsymbol{V}$ based on a subset ${\boldsymbol{X}}^{(i)}_{\text{o}}$ of the remaining attributes $\boldsymbol{X}^{(i)}=\boldsymbol{V} \setminus \boldsymbol{Y}^{(i)}$. Note that ${\boldsymbol{X}}^{(i)}_{\text{o}}$ can be either all the observed attributes under $\boldsymbol{X}^{(i)}$, or specified by the enquirer who wishes to know the impact of only certain evidence 
${\boldsymbol{X}}^{(i)}_{\text{o}}\subseteq \boldsymbol{X}^{(i)}$
on the targets. In this case, the conditional probability to be estimated for the $i$-th data instance becomes $P_{\boldsymbol{\Theta}}(\boldsymbol{Y}^{(i)}|{\boldsymbol{X}}^{(i)}_{\text{o}})$ where $\boldsymbol{\Theta}$ stands for the sets of parameters for all the possible target attributes across the (training) data instances. 

Naively, one could leverage reduced-feature modelling~\citep{saar2007handling} for handling the above task. More specifically, a composite of discriminative models can be trained, each dedicated to a specific attribute $X_j \in \boldsymbol{V}$ as the target, based on every possible subset of $\boldsymbol{V} \setminus \{ X_j\}$. However, this training strategy could become infeasible over only a dozen attributes. Alternatively, one could learn a new composite model whenever predictions for a new/test instance is required~\citep{Friedman1996lazy}. In this case, the model is learned from a training dataset that only contains the attribute subset ${\boldsymbol{X}}^{(i)}_{\text{o}}$ of the $i$-th new instance. This strategy, however, significantly increases the prediction time, which is undesirable for real-world applications.

Another alternative is to train a model to predict each $x_j$ based on all the remaining features $\boldsymbol{V} \setminus \{X_j\}$. At the testing stage, the instance-specific targets can be computed simultaneously as $P_{\boldsymbol{\Theta}}(\boldsymbol{Y}^{(i)}|{\boldsymbol{X}}^{(i)}_{\text{o}}) \propto \prod_{j} P_{\boldsymbol{\theta}_j}(Y^{(i)}_{j}|{\boldsymbol{X}}^{(i)}_{\text{o}})$. All the ignored and/or missing attributes ${\boldsymbol{X}}^{(i)}_{\text{m}}=\boldsymbol{X}^{(i)} \setminus {\boldsymbol{X}}^{(i)}_{\text{o}}$ can have their values masked by zeros if they are binary attributes, or set to a distinct category (e.g. to represent the values being missing). Such treatments also need to be applied to the training stage to allow for consistency in the conditional probability estimation.   




The above treatments, however, have two major issues. First, both masking and having a separate category can yield biased estimates for the conditional probabilities. Moreover, for continuous attributes, the masking is not principled as zeros have numerical meanings. A second issue is that the independence assumption for the discriminative models learned for each target is not effective in many real-world applications. For example, in NLP tasks, it is ideal to predict target words under autoregressive modelling rather than independence modelling to capture strong dependencies between them (e.g.\ ``I went to \textit{New York} to see the \textit{Empire State building}" where the words ``New York" and ``Empire State building" have strong dependencies). 

Conditional random fields (CRFs)~\citep{Lafferty2001crfs} provide solutions to the above issue. They model the inter-dependencies between a predefined set of targets with various structures, e.g. linear chain CRFs~\citep{sutton2012introduction}, and estimate their conditional probabilities with efficient inference techniques (e.g. the forward-backward algorithm). However, since CRF classification has to pre-define the targets and therefore the input attributes as well, they are not suitable for the target agnostic prediction task.


\subsection{Generative Learning}\label{sec:generative_learning}

\sloppy Generative classification models focus on inferring the joint distribution $P(Y,\boldsymbol{X})$ of all the attributes involved in a problem in order to calculate the conditional probability $P(Y|\boldsymbol{X})$. They achieve this by leveraging 
Bayesian networks~\citep{jensen1996introduction} for efficient factorization of the joint distribution into simpler (probabilistic) functions. For example, naive Bayes is the most common generative learning model which factorizes the joint distribution into a product $P_{\boldsymbol{\theta}_{0}}(Y)\prod_{j} P_{\boldsymbol{\theta}_j}(X_j|Y)$ and further employs Bayes rules to calculate $P(Y|\boldsymbol{X})$. 

Overall, the factorization allows for tractable estimation of the function parameters (e.g. marginal and conditional probability parameters $\boldsymbol{\theta}_0, \boldsymbol{\theta}_j$ in Naive Bayes). Traditionally, the parameters of a generative classification model are estimated by maximizing the log-likelihood with regularization to allow the classifier to best fit the data at hand and generalise beyond it. Recently, more sophisticated classifiers have been proposed based on Bayesian networks~\citep{martinez2016scalable,zaidi2017efficient,petitjean2018accurate}. They perform more accurate prediction by effectively modelling inter-dependencies between attributes and learning their associated parameters via either tailored regularisation \citep{petitjean2018accurate} or additional discriminative objectives \citep{martinez2016scalable,zaidi2017efficient}. However, all these models are specifically designed for predicting a single predefined target attribute and therefore, cannot be directly applied to target-agnostic prediction.

Generalized from the generative classification models, generative graphical models~\citep{koller2009probabilistic} aim to infer the joint distribution with no particular target attribute. They are capable of capturing inter-dependencies between attributes and possess moderately efficient inference algorithms which enable marginalization of the joint distribution over any set of the attributes. As a result, generative graphical models support target-agnostic learning. However, the flexibility in their predictions is often acquired at the cost of prediction accuracy~\citep{ng2001comparison}. Furthermore, it is usually very difficult to scale the marginalization of graphical models towards arbitrary multiple (and potentially thousands of) targets~\citep{petitjean2013scaling}. Recently, progress has been made in leveraging the decomposibility of chordal graphs to scale up the construction of corresponding graphical models such as markov random fields and junction trees to thousands of attributes in seconds instead of days~\citep{PetitjeanWebb15}. However, further scalability still remains an open question that requires more research attention.

\subsection{Self-Supervised Learning}

Self-supervised learning~\citep{desa1993learning} aims to learn useful representations/embeddings for data instances using (deep) neural networks created and pre-trained based on well-selected artificial supervised learning tasks. 
It has become a standard tool to achieve state-of-the-art performance in many tasks of computer vision \citep{jing2020self} and natural language processing \citep{qiu2020pretrained}. The learning of the representations can be conducted under the paradigm of generative learning where the networks estimate the joint distribution of the non-target variables $P(\boldsymbol{X})$; which is commonly referred to as the ``pre-training'' of the networks. Then, the learned representations can be deployed via transfer learning \citep{zhang2020text} to boost the performance of discriminative prediction $P(Y|\boldsymbol{X})$ where the single target variable $Y$ has too few labels to perform any reliable supervised learning. More specifically, the joint distribution estimation of the networks is performed with a general prediction formula:
\begin{equation}
  \mathbb{E}_{\widetilde{\boldsymbol{X}}\sim q(\boldsymbol{X})}\big[P_{\boldsymbol{\Theta}}(\boldsymbol{X}|\widetilde{\boldsymbol{X}})\big]
  \label{eqn:self_supervised_define}
\end{equation}
where $q(\cdot)$ is a randomised data modification
operator that may mask, add noise to or permute 
a selection of attributes in $\boldsymbol{X}$; $\boldsymbol{\Theta}$ are the sets of network parameters for each attribute over the training data. For the $i$-th training data instance $\boldsymbol{X}^{(i)}$, its modified version $\widetilde{\boldsymbol{X}}^{(i)}\sim q(\boldsymbol{X}^{(i)})$ and in this case, part of $\boldsymbol{X}^{(i)}$ will be unmodified; that is $\widetilde{\boldsymbol{X}}^{(i)}=[\widetilde{{\boldsymbol{X}}}^{(i)}_{\text{m}};{\boldsymbol{X}}^{(i)}_{\text{o}}]$.
Here 
$\boldsymbol{X}_{\text{o}}^{(i)}$ represents 
unmodified or original observed attributes, and
$\widetilde{\boldsymbol{X}}^{(i)}_{\text{m}}$ represents the  attributes that are corrupted or transformed by $q(\cdot)$.
Pre-training is usually performed as the task of maximising the model's performance at the prediction specified by Equation (\ref{eqn:self_supervised_define}), which is determined by the data operator $q(\cdot)$.

In general, the probability model $P_{\boldsymbol{\Theta}}(\boldsymbol{X}|\widetilde{\boldsymbol{X}}_{\text{m}}, {\boldsymbol{X}}_{\text{o}})$ described above
represents how the original data is modelled and recovered based on the modified and the original parts of the  data. 
It can be modelled by various types of deep neural networks such as DAEs~\citep{vincent2008extracting}, GANs~\citep{Goodfellow2014gan}, Auto-regressive (AR) Neural Networks etc. After the pre-training phase, the networks will be ``fine-tuned'' by reusing the learned but trainable representations from its certain hidden layers to predict a new single target $y$ via transfer learning. The assumption is that this downstream task and the pre-training task should have similarity in terms of ``lower-level'' representations.

The latest significant performance improvement brought by self-supervised learning in NLP and computer vision tasks is from BERT, which is pre-trained on the following cost function based on a masked language model:
\begin{equation}
   \sum^{n}_{i=1}\sum_{j:\widetilde{X}^{(i)}_j\in \widetilde{\boldsymbol{X}}^{(i)}_m} 
   \log P_{\boldsymbol{\Theta}}\left(X^{(i)}_{j}~|~
   \widetilde{\boldsymbol{X}}^{(i)} \right)\text{;}~~~~\widetilde{\boldsymbol{X}}^{(i)}\sim mask(\boldsymbol{X}^{(i)}) \label{eqn:self_supervised_define0}
\end{equation}
where the $mask(\cdot)$ function, which is an instance of $q(\cdot)$ in Equation~(\ref{eqn:self_supervised_define}), randomly masks a small percentage of the input attributes. From Equation~(\ref{eqn:self_supervised_define0}), the $j$-th attribute of the $i$-th training data instance is masked as $\widetilde{X}^{(i)}_j$ and therefore, it belongs to the masked attribute set $\widetilde{\boldsymbol{X}}^{(i)}_m$. Its original value is then recovered based on $\widetilde{\boldsymbol{X}}^{(i)}$ which includes both the masked and unmasked attributes. It is clear that Equation~(\ref{eqn:self_supervised_define0}) is a variation of Equation~(\ref{eqn:self_supervised_define}).



Recently, advances in contrastive learning~\citep{JMLR:v13:gutmann12a,hjelm2018learning,oord2018representation} have provided another path to achieve self-supervised learning under the paradigm of discriminative learning. Broadly speaking, contrastive learning aims to ``learn to discriminate'' positive samples of the data from negative samples by performing a non-linear logistic regression~\citep{JMLR:v13:gutmann12a} or mutual information maximization~\citep{oord2018representation} over the representations instead of using them to reconstruct the data as in the generative case.


\subsection{pseudo-likelihood theory}

The idea of self-supervised learning can also be viewed from the perspective of pseudo-likelihood theory~\citep{besag1975statistical}. This theory provides an efficient approximation to the joint distribution which is usually impractical to be calculated exactly due to a complex normalization constant. It avoids directly computing this term by factorizing the joint distribution into a product of univariate conditional distributions of each attribute given all the other attributes. As a result, it corresponds to the self-supervised training specified by Equation~(\ref{eqn:self_supervised_define0}) without the masking operator:
\begin{equation}
    P(\boldsymbol{X})\approx\prod_{X_j\in \boldsymbol{V}} 
    P_{\boldsymbol{\Theta}}(X_{j}~|~\boldsymbol{X}\setminus \{X_{j}\})~\label{eqn:self_supervised_define1}
\end{equation}
In Equation~(\ref{eqn:self_supervised_define1}), $P_{\boldsymbol{\Theta}}(X_{j}~|~\boldsymbol{X}\setminus \{X_{j}\})$ is the conditional distribution of the $j$-th attribute given all the other attributes. This distribution can also be viewed as a function $f_{j}(\boldsymbol{X}\setminus \{X_{j}\};\boldsymbol{\Theta})$ which predicts $X_j$ based on the remaining attributes $\boldsymbol{X} \setminus \{X_j\}$ with parameter sets $\boldsymbol{\Theta}$. For all the conditional distributions, they can be modelled by the same function above with shared $\boldsymbol{\Theta}$. For self-supervised learning, it can be described as learning such a shared model $\boldsymbol{f}(\boldsymbol{X}; \boldsymbol{\Theta})$ for a task of predicting any parts of its input using any observed part. In general, the learning can be done by fitting the function across every attribute (for every training instance): $\prod\nolimits_{j}f_{j}(\boldsymbol{X}\setminus \{X_{j}\};\boldsymbol{\Theta})$ or by fitting it stochastically as $\mathbbm{E}_{j\sim\text{Uniform}}[f_{j}(\boldsymbol{X}\setminus \{X_{j}\};\boldsymbol{\Theta})]$ with respect to the shared $\boldsymbol{\Theta}$. The stochastic learning strategy has been adopted (and adapted) by most self-supervised learning models (e.g. as the masked language modelling by BERT) as it is much less computationally expensive when the dimension $|\boldsymbol{X}|$ is large.

Past research has focused on proving properties (e.g. consistency and uniqueness) of the pseudo-likelihood theory on specialised distributions and probabilistic models in various applications~\citep{pensar2017marginal}. However, in the context of self-supervised learning which typically has much more general models, these properties remain largely unproved. Below, we provide the following general theorem: that pseudo-likelihood is as good as likelihood for large data
when the true model lies within the model family.

\begin{thm}
 Let $\boldsymbol{X}$ be a set of attributes, and let $g(\boldsymbol{X};\boldsymbol{\Theta})$ be a full joint probability model for 
$\boldsymbol{X}$ with parameterization $\boldsymbol{\Theta}$ and with no zeros
(i.e., $g(\boldsymbol{X};\boldsymbol{\Theta})>0$ for all $\boldsymbol{X}$ and $\boldsymbol{\Theta}$).
Furthermore, let the model family be indentifiable. 
Consider $n$ IID data  $\boldsymbol{X}^{(1)},...,\boldsymbol{X}^{(n)}$ for some ground-truth parameters $\boldsymbol{\Theta}_*$.
Then the parameters maximising the pseudo-likelihood function for the data
will converge as $n\rightarrow \infty$ 
 to the ground-truth parameters $\boldsymbol{\Theta}_*$, 
 as do the parameters maximising the likelihood function.
\end{thm}
\noindent
\paragraph{Outline of proof:} The pseudo-likelihood represents a product of a set of conditional likelihoods, for each attribute $X_j \in \boldsymbol{X}$, given by 
$\prod_{i=1}^n P\big(X^{(i)}_j\, |\, \boldsymbol{X}^{(i)}\setminus \{X^{(i)}_j\}, g(\cdot;\boldsymbol{\Theta})\big) $, where $X^{(i)}_j$ denotes the value of attribute $X_j$ at the $i$-th data instance. For infinite data, $\boldsymbol{\Theta}_*$ is a maximum for this product and therefore, provides one solution in the limit. However, it is possible that other distinct parameterizations,
$\boldsymbol{\Theta}'$, also satisfy all the conditional maximum likelihoods. Now, it is not generally known that the Hammersley-Clifford theorem applies to more general cases than those on discrete spaces, but the proof on Wikipedia\footnote{\tt https://en.wikipedia.org/wiki/Hammersley\%E2\%80\%93Clifford\_theorem} applies generally to probability measures. By this more general Hammersley-Clifford theorem, which can apply because there
are no zeroes, only one distribution can be consistent with all the conditionals, and since $\boldsymbol{\Theta}_*$ satisfies this property, no other distinct conditional maximising distribution can exist.~\rule{2mm}{2mm} \\[5pt]

Note this theorem relies on two strong conditions of our model family:
(1)  ``truth" is within the model family, and (2) the model family is identifiable. We argue that with sufficiently complex deep neural networks, the first is a reasonable assumption to make. Moreover, the second is really for technical convenience, so is arguably not an issue. Nonetheless, the theorem shows us that pseudo-likelihood training of a network is able to converge on an implicit full joint model of the data, even though training only works with each individual conditional. In other words, appropriate discriminative training can fit a generative model. We believe this provides one theoretical interpretation for some versions of self-supervised training. It is also a comparable but rather simpler and better justification than the information-theoretic interpretation of \citet{kong2019mutual}.

\section{Target-Agnostic Learning}
Target-agnostic learning is a general machine learning paradigm that brings together the merits of discriminative learning, generative learning and self-supervised learning. In this paper, we give it the following formal definition: 

\begin{defn} 
\textit{A target-agnostic learning system learns a model from the data described by a fixed set of attributes $\boldsymbol{V}$ and is able to predict for any new data instance the values of its arbitrary target attributes $\boldsymbol{Y} \subset \boldsymbol{V}$ from any subset $\boldsymbol{X}$ of the remaining attributes $\boldsymbol{X} \subseteq \boldsymbol{V} \setminus \boldsymbol{Y}$.}
\end{defn}

Note that in this definition,
prediction matches the training style of some versions of 
self-supervised learning \citep{qiu2020pretrained}, where masking is conducted 
(i.e. $\boldsymbol{X} \subset \boldsymbol{V} \setminus \boldsymbol{Y}$), or
pseudo-likelihood training where no masking is done
(i.e. $\boldsymbol{X} = \boldsymbol{V} \setminus \boldsymbol{Y}$).
Moreover, target-agnostic learning subsumes missing value imputation and fully generative graphical models as mentioned before. In~the next section, we will expand on these relationships and discuss some corresponding standard algorithms for target-agnostic learning.

\subsection{Special Case: Missing Value Imputation}

It follows from Definition 1 that missing value imputation is a special case of target-agnostic learning. It assumes that multiple missing attributes need to be imputed for some data instances before model training or prediction. There are some subtle differences in the ways the imputation is conducted for each of these scenarios; but both are special cases of target-agnostic learning.

The main idea of missing value imputation at training time is to replace each missing value with an estimate to complete the ``training'' instances from which a model can then be learned to predict a pre-specified target attribute. The estimation process in this case exhibits a trade-off between the computation time and the estimation accuracy. Simple imputation techniques such as the most frequent value imputation for categorical attributes are extremely fast but in general perform poorly at prediction. Sophisticated techniques such as random forest based imputation~\citep{stekhoven2012missforest} and EM based imputation~\citep{graham2009missing} require building separate models that iteratively impute the missing attributes based on the observed ones (some being more complicated by iteratively imputing each attribute with the currently imputed values of all the other attributes). 

Apart from the above point estimation approaches, missing value imputation can also be conducted with sampling distributions, which is referred to as multiple imputation~\citep{graham2007many}. More specifically, multiple versions of a training instance are generated by filling each missing attributes with values drawn from their respective posterior distributions given the observed attributes. The ``pseudo'' training instances are then passed on to train different versions of the models that predict the same target attribute. At the prediction time, multiple predictions are drawn from the trained models and then aggregated in certain fashion to obtain the final prediction for the target attribute. 

As for missing value imputation at the prediction time, similar ideas can be applied \citep{saar2007handling}. In this case, each missing value within a ``test'' instance is estimated before a trained model can be applied to predict the pre-specified target. Alternatively, multiple imputation followed by prediction aggregation can also be leveraged for capturing the uncertainty within the estimation. Its difference from the training-time missing value imputation is that the imputation models for prediction time can be built on both the test instances and the previous training instances. However, building such models will become very inefficient when test instances come in a stream as every time a new model has to be built that predicts on a different set of missing attributes. 

On the other hand, reduced-feature modelling \citep{saar2007handling}, designed specifically for missing value imputation at prediction time, aims to train a different model from a data subset that only contains the observed attributes of a test instance. Then, it uses this model to solely predict the target attribute of this particular instance. Three variants are conceivable in this case: (1) learning all the possible models before prediction time, which is infeasible beyond a dozen variable; (2) learning a new model at prediction time, known as ``lazy classification''~\citep{Friedman1996lazy}, which is infeasible for most applications because of the time required to perform prediction; and (3) marginalising over the observed attributes of a learned model, which is generally feasible with simple models (that have limited predictive capacity) or with a few attributes\footnote{In the general case, marginalizing is exponential with the number of attributes over which the marginalization is performed.}.

Most missing value imputation techniques for prediction time build discriminative and ad-hoc models to deal with the different sets of observed and missing attributes encountered in test datasets. In this case, the discriminative imputation models built for training time cannot be directly applied (e.g. if there exist missing attributes at prediction time that are not missing at training time), and vice versa. As a result, neither the training-time imputation techniques nor the prediction-time techniques can comprehensively address the missing value problem. In comparison, target-agnostic learning is inherently generative and provides principled approaches to learn unified fully-joint models for missing value imputation at both training and prediction time. These models apply principled marginalization techniques to readily predict for arbitrary missing attributes of either training or test instances given their arbitrary observed input attributes. 

\subsection{A Generative Graphical Model Learning Approach for Target-Agnostic Learning}

As mentioned earlier in Section~\ref{sec:generative_learning}, generative graphical models aim to learn the joint distribution with no particular target and can perform marginalization over any set of unobserved attributes to predict for multiple arbitrary targets. Recent research advancements have made significant progress in addressing the main issues of graphical models, including the scalability of learning the model structures and the efficiency and accuracy of marginalization algorithms~\citep{PetitjeanEtAl14a,PetitjeanWebb15, PetitjeanWebbTut16}. Therefore, it seems both reasonable and promising to base target-agnostic learning on generative graphical models.

Chordalysis
\citep{petitjean2013scaling,PetitjeanEtAl14a,capdevila2018experiments} is a state-of-the-art graphical model learning framework. It is able to scale the learning of the generative graphical models from discrete data up to thousands of attributes. The scalability comes from an efficient forward selection strategy for selecting a new best model (structure) from a set of candidates to replace the previous best one (at each iteration of the selection process). The efficiency of the selection strategy relies on the decomposability of chordal graphs which accelerates the search process of graph edges incrementally added to the structure. The chordal graph models also possess an important property that the graph of their maximal cliques is a tree, which is also known as the junction tree. A merit of junction tree models is that they allow for efficient and exact marginalization over any set of attributes. 

Furthermore, Chordalysis has employed advanced scoring functions for selecting statistically significant decomposable models at each iteration. They include the Subfamilywise Multple Testing (SMT) score \citep{webb2016multiple} and the Quotient Normalised Maximum Likelihood (QNML)~\citep{pmlr-v84-silander18a}. As for estimating the parameters (i.e. conditional probability tables) of the learned model structure, Chordalysis originally employed maximum likelihood estimation with M-estimation smoothing~\citep{Zadrozny2001}. Recently, smoothing techniques based on hierarchical Dirichlet Processes have been proposed and shown to exhibit better regularization effects on the parameter estimation for graphical models~\citep{petitjean2018accurate,zhang2020bayesian}.

\subsection{Self-supervised Learning Approaches for Target-Agnostic Learning}
Self-supervised learning is a family of deep learning approaches that leverage target-agnostic learning for model induction. More specifically, during each training epoch, a random subset of the attributes of each training instance is modified to become the targets by either masking, noise addition or permutation. To learn to recover the modified values, self-supervised learning models maximize the pseudo-likelihood of the target attributes conditioned on the rest of the input across the training instances as specified in Equation (\ref{eqn:self_supervised_define0}).

There exist different types of self-supervised learning models which have been inducted based on the above target-agnostic learning paradigm. They include the autoencoder models, the generative adversarial models and the masked language models. In this paper, we will study the most representative instances of these different types of models in terms of their target-agnostic learning performance.

\subsubsection{Autoencoder Models}
The goal of the autoencoder (AE) models is to learn useful latent representations of an input by reconstructing (part of) the input based on (the remaining part of) itself. Thanks to the simplicity and flexibility of the AE models, they have served as the foundation of self-supervised learning. Here, we present two popular types of AE models: denoising autoencoders (DAE) and variational autoencoders (VAE)~\citep{kingma2013auto}, both of which have been shown to be able to perform target-agnostic learning for model induction. 

The main idea behind DAE models is that the representations should be learned in such a way that they are insensitive to noise-corrupted inputs (at both training and prediction time). To achieve this, during the training time, the original input $\boldsymbol{X}$ is modified to become $\widetilde{\boldsymbol{X}}$ by masking or adding noise to a random part of the input. The modified input $\widetilde{\boldsymbol{X}}$ is then mapped through some feed-forward neural network (FNN), known as the encoder, into a latent representation $\boldsymbol{Z}$. Based on this representation, the models will learn to reconstruct the original input $\boldsymbol{X}$ using a corresponding decoder FNN. The above training process can also be formulated as the approximation of the joint distribution over $\boldsymbol{X}$ as follows:
\begin{equation}
  \mathbb{E}_{\widetilde{\boldsymbol{X}}\sim q(\boldsymbol{X})}\mathbb{E}_{\boldsymbol{Z}\sim P_{\boldsymbol{\Theta}'}(\boldsymbol{Z}|\widetilde{\boldsymbol{X}})}\big[P_{\boldsymbol{\Theta}}(\boldsymbol{X}|\boldsymbol{Z})\big]
  \label{eqn:self_supervised_define2}
\end{equation}
where $\boldsymbol{Z}$ is the latent representation of $\boldsymbol{X}$ and is generated from the distribution $P_{\boldsymbol{\Theta}'}(\boldsymbol{Z}|\widetilde{\boldsymbol{X}})$, which is modelled by the encoder network with a set of parameters $\boldsymbol{\Theta}'$.

\citet{gondara2018mida} first proposed to apply the classifical DAE to missing value imputation, which they refer to as \textbf{MIDAS}. For each training instance, MIDAS randomly masks a certain percentage of its attributes by zeroing their values and further dropouts certain percentages of the hidden neuron outputs throughout the fully-connected hidden layers. It then attempts to reconstruct the masked attributes for each instance to learn robust representations for the data by maximizing the approximated joint distribution specified in Equation~(\ref{eqn:self_supervised_define2}). 

In the original paper, the DAE network is trained multiple times with different initial values for the network weights. At the prediction time, a ``test'' data instance is fed to the different trained DAE networks with the missing values of its target attributes replaced by zero. These DAE networks then yield different predictions for each target attribute of this instance. The predictions will be aggregated (i.e. by averaging for continuous attributes and by counting the most frequent value for categorical attributes) to yield the final prediction. However, the overhead of training the DAE network multiple times can be huge with large-scale data. Alternatively, in this study, we train the DAE network only once and the multiple predictions for each target attribute can still be obtained due to the stochasticity induced by the dropout operations across the hidden layers.

Unlike the DAE models which focus on recovering corrupted inputs, VAE models aim to approximate intractable posterior distributions of latent representations $\boldsymbol{Z}$ for generative purposes. Therefore, they are not originally designed with random modification on the inputs. Recently, \citet{ivanov2018variational} has integrated VAE with the random masking of the inputs for learning a joint model of $\boldsymbol{X}$, which is referred to as \textbf{VAEAC}. The idea is to approximate the joint distribution by learning all the conditional distributions on each attribute instead. The approximation is achieved by extending the conditional variational autoencoder (CVAE) \citep{sohn2015conditional} framework with the idea of target-agnostic learning: that is to arbitrarily partition the input (from each training instance) into observed and masked attributes in every training epoch; such that the conditional posterior of each unobserved attribute is a function to be learnt on the observed attributes as follows:
\begin{equation}
  \mathbb{E}_{\widetilde{\boldsymbol{X}}\sim q(\boldsymbol{X})}\mathbb{E}_{\boldsymbol{Z}\sim P_{\boldsymbol{\Theta}'}(\boldsymbol{Z}|\boldsymbol{X}_\text{o},\boldsymbol{M})}\left[\prod_{j:\widetilde{X}_j\in \widetilde{\boldsymbol{X}}_m}P_{\boldsymbol{\Theta}}(X_{j}|\boldsymbol{Z},\boldsymbol{X}_\text{o}, \boldsymbol{M})\right]
  \label{eqn:self_supervised_define3}
\end{equation}
where $\boldsymbol{M}$ is a set of binary variables that correspond to each attribute in $\widetilde{\boldsymbol{X}}$ and indicate which of them is unobserved/masked (for each training instance at every training epoch). It takes on value 1 if the attribute is masked and 0 otherwise. 

In Equation~(\ref{eqn:self_supervised_define3}), the posterior distribution of latent representations $\boldsymbol{Z}$ (which corresponds to the VAE encoder) is modelled to be dependent on both the observed attributes $\boldsymbol{X}_{\text{o}}$ and the binary mask vector $\boldsymbol{M}$. The decoder is set to maximize the pseudo-likelihood of the masked attributes $\widetilde{\boldsymbol{X}}_{\text{m}}$ conditioned on simultaneously $\boldsymbol{Z}$, $\boldsymbol{X}_{\text{o}}$ and $\boldsymbol{M}$. Same as the MIDAS model, the values of the masked attributes in this case are zeroed in the input during both the training and the prediction time. For prediction, considering that the latent representations $\boldsymbol{Z}$ are stochastic: the encoder outputs distributional parameters to $P_{\boldsymbol{\Theta}'}(\boldsymbol{Z}|\boldsymbol{X}_\text{o},\boldsymbol{M})$, we thus adopt the multiple imputation strategy (based on sampled $\boldsymbol{Z}$ from the trained encoder over each test instance) for predicting each target attribute.

Apart from the random masking, modification by randomly permuting the input into different orderings has also been applied by the target-agnostic learning approaches. More specifically, \citet{germain2015made} proposed \textbf{MADE}, a permutation-based autoencoder. It imposes an autoregressive constraint on the joint model learning: the conditional probability of each attribute can only be based on the previous ones in a given ordering of the attributes. This constraint can be expressed by the following formula on decomposing the joint distribution into a product of conditionals:
\begin{equation}
    P(\boldsymbol{X})\approx\prod^{J}_{j=1}P(X_j|\boldsymbol{X}_{<j})
\end{equation}
where $\boldsymbol{X}_{<j}$ represents the set of attributes that precede attribute $X_j$ in the particular ordering. In each training epoch, random orderings of the attributes of each training instance are generated and passed through an autoencoder to estimate the joint distribution with the following prediction formula: 
\begin{equation}
  \mathbb{E}_{\boldsymbol{t}\sim q(J)}\left[\prod^{J}_{j=1} P_{\boldsymbol{\Theta}}(X_{t_j}|\boldsymbol{X}_{<t_j})\right]
  \label{eqn:self_supervised_define4}
\end{equation}
where $\boldsymbol{t}$ is an ordering vector with permuted attribute IDs generated by the randomised permutation function $q(\cdot)$ on indices. Correspondingly, a masking strategy is employed to remove connections between network layers (i.e. by element-wise multiplication of each weight matrix with respective binary mask matrices) to respect the autoregressive constraint imposed by the generated ordering. 

At the prediction time, we employ multiple imputation based on different orderings of the observed attributes for predicting each target attribute. To achieve this, we can place the target attributes (with zero values) subsequent to the observed attributes in the randomised orderings. This ensures that the target attributes can be predicted based on the observed ones. Alternatively, we can adopt masked language modelling for both the training and prediction time; that is randomly masking a small percentage of the input values given their specific input orderings, followed by 
the reconstruction of the masked values based on the observed attributes (and the masks). In the experiments, we have not found any notable performance difference between these two strategies.


\subsubsection{Generative Adversarial Networks}
Unlike the autoencoder models which primarily adopt a generative learning paradigm, generative adversarial networks (GANs)~\citep{Goodfellow2014gan} combine the paradigms of both generative and discriminative learning for performing target-agnostic learning. Recently, \citet{yoon18a} has proposed an extension on GANs for missing value imputation, referred to as \textbf{GAIN}. It inherits the generator-discriminator architecture from GANs. During the training time, the generator aims to impute the unobserved attributes of the training instances conditioned on their observed attributes to complete the entire instances. To adapt their model to target-agnostic learning, we randomised the masking of each training instance and had the generator impute the random target attributes at every training epoch. The imputation can be viewed as learning a conditional model with the following prediction formula:
\begin{equation}
  \mathbb{E}_{\widetilde{\boldsymbol{X}}\sim q(\boldsymbol{X})}\mathbb{E}_{\boldsymbol{Z}\sim \text{Unif}}\left[\prod_{j:\widetilde{X}_j\in \widetilde{\boldsymbol{X}}_m}P_{\boldsymbol{\Theta}}({X}_j|\boldsymbol{Z}, \widetilde{\boldsymbol{X}},\boldsymbol{M})\right]
  \label{eqn:self_supervised_define5}
\end{equation}
where $\boldsymbol{Z}$ is a set of uniformly distributed noise variables sampled between 0 and 0.01, and it corresponds to the set of modified attributes $\widetilde{\boldsymbol{X}}$. In the paper, the above formula is specifically implemented as a feed-forward neural network $g(\cdot;\boldsymbol{\Theta})$ to predict for the masked attributes:
\begin{equation}
    \hat{\boldsymbol{X}}_{m}=g\big((\boldsymbol{1}-\boldsymbol{M})\odot\widetilde{\boldsymbol{X}}+\boldsymbol{M}\odot\boldsymbol{Z};\boldsymbol{\Theta}\big)
\end{equation}
where $\hat{\boldsymbol{X}}_m$ is the set of predicted attributes corresponding to the set of masked attributes $\widetilde{\boldsymbol{X}}_m$. Based on the imputation $\hat{\boldsymbol{X}}=[\hat{\boldsymbol{X}}_\text{m};\boldsymbol{X}_{\text{o}}]$, the discriminator will learn to distinguish the observed attributes of each instance from the imputed ones. More specifically, it takes in the imputation $\hat{\boldsymbol{X}}$ and a set of hint binary variables $\boldsymbol{H}$ which randomly reveals a percentage of the masked attributes $\widetilde{\boldsymbol{X}}_m$ of each instance to the discriminator. The hint variables correspond to the mask variables $\boldsymbol{M}$ and achieve the goal by $\boldsymbol{M}\odot\boldsymbol{H}$. In other words, they act as the ``mask variabes'' for $\boldsymbol{M}$ which are now the targets in the discriminator. Finally, the discriminator is implemented as the following network $f\big([\hat{\boldsymbol{X}};\boldsymbol{M}\odot\boldsymbol{H}];\boldsymbol{\Theta}'\big)$: $\mathbbm{R}^{2J}\rightarrow[0,1]^{J}$ which predicts the probabilities of the attributes being masked.

The GAIN model is also characterized by its loss function which comprises the discriminator loss and the generator loss. For the discriminator loss, it is the binary cross entropy loss over the values of mask attributes $\boldsymbol{M}$ against the predicted values from the network $f(\cdot;\boldsymbol{\Theta}')$ for all the training instances. As for the generator loss, it further consists of two component losses. The first one ensures that the generated imputation $\hat{\boldsymbol{X}}_{\text{m}}$ successfully fools the discriminator (as defined by the minimax game) to make it believe that they are observed attributes. It is the log-loss of the actual mask values\footnote{The actual mask values in this case are all ones.}, which correspond to the masked attributes across the training data, against their predicted values from the discriminator. The second one ensures that the generator accurately predicts the values of all the observed attributes and hence, is the reconstruction loss on the observed training data. The two component losses are balanced by a trade-off hyper-parameter attached to the reconstruction loss. At the prediction time, we use the trained generator of the GAIN model to perform multiple imputation for predicting each target attribute across the test data.


\subsubsection{Transformer Language Models}
\textbf{BERT}~\citep{devlin2018bert} is a deep natural language model originally proposed for pre-training contextual representations of word tokens under a target-agnostic learning paradigm. During the pre-training, it randomly masks a certain percentage of the input tokens (e.g. in a pair of sentences) by replacing them with a ``[MASK]'' token. Then, BERT attempts to recover the masked tokens using stacks of fully-connected transformers over the embedding layer for each input token (i.e. both the observed and the masked tokens). The objective function, which is a pseudo-likelihood approximation of the joint distribution, of this target-agnostic prediction task has been previously specified in Equation~(\ref{eqn:self_supervised_define0}). Other than the main prediction task, traditional BERT also employs a side prediction task (e.g. next sentence prediction for sentence pairs) which helps the learning of the representations of each token.

To use BERT in our study, we treat every possible value of each attribute as a unique ``token'' and perform the target-agnostic learning to train the BERT model to learn representations for each attribute value. When an attribute is continuous, we convert its values into discrete ones using equal-frequency discretization. Note that in this case, each output of BERT corresponds to a specific attribute and therefore, only the (masked) attribute's own values should be considered (among the entire ``vocabulary'' of attribute values) when it is being predicted. To achieve this, we have added a masking layer on top of the final softmax layer of BERT to set the probabilities of the irrelevant values to zero for each attribute. Then, the masked probabilities are passed to a re-normalization layer in order to obtain the new softmax probabilities. Furthermore, in our study, we have removed the side prediction task (and its corresponding ``[CLS]'' token) from BERT to make it solely rely on the target-agnostic prediction task. At the prediction time, the target attributes are input with the ``[MASK]'' tokens into the trained BERT model to have their values predicted.  

\textbf{XLNet}~\citep{yang2019xlnet} is another deep natural language model which has been shown to outperform BERT in a wide range of natural language processing tasks. It adopts the random (permutation-based) autoregressive modelling from MADE~\citep{germain2015made}, as specified by Equation~(\ref{eqn:self_supervised_define4}), to perform the target-agnostic learning and combines it with a stacked transformer architecture. It overcomes the main weakness of BERT which is the assumption of the independence of the masked tokens and meanwhile, preserves the modelling of the bidirectional contexts of the tokens. In addition, XLNet chooses to stack up the Transformer-XL \citep{dai2019transformer} encoders to build up its architecture. These encoders are characterised by their memory caching mechanism which allows the intermediate embeddings from the encoders of preceding tokens in the orderings to be memorised and affect the encoder outputs of the current token. The length of the memory cache decides how many previous tokens can affect the prediction of the current one.

Following our training settings for the BERT model, we treat every possible value of each attribute as a unique ``token'' for XLNet as well. We adopt the same test settings for XLNet; the test datasets contain the ``[MASK]'' tokens for each instance which represent the missing values for their unobserved attributes. Correspondingly, we introduce the ``[MASK]'' tokens into the training of the XLNet model. More specifically, in each training epoch, we randomly mask a certain percentage of the input attributes of each instance and then present them to the XLNet model in random orderings. Different from BERT (and the aforementioned alternative training strategy for MADE) which solely focuses on recovering the masked values, XLNet recovers the values of both the observed and masked attributes. This is attributed to its unique two-stream self-attention mechanism which ensures that when being predicted, an observed attribute cannot ``see'' its ground-truth value from the input layer, but its information can still flow to the rest part of the stacked transformer architecture to help predict the values of all the other attributes. At the prediction time, multiple imputations are generated based on different random orderings of the tokens (including the ``[MASK]'' token) in a test instance and are averaged to obtain the final prediction for each target attribute of the instance.

\begin{table}[t]
\centering
  \begin{tabular}{llcccccc}
    \toprule
     \textbf{Data Type}& \textbf{Data} & \textbf{\#Instance}& \textbf{\#Attribute} & \textbf{\#Test Target (20\%)}\\
    \midrule
     &Mushroom&8,124&23&6,887\\
     &Chess&28,056&7&7,798\\
     &EPESE&14,456&27&13,372\\
     &Coil&9,000&86&33,871\\
          &Adult&48,842&15&28,921\\
     \textbf{Categorical}&Telecom\_Churn&51,047&22&44,841\\
     &Finance&3,453&497&64,828\\
     &Connect-4&67,557&43&115,951\\
     &Fars&100,968&29&121,322\\
     &Kaggle\_Survey&23,859&346&329,465\\
     &Eron\_Email&5,172&3,001&619,489\\
     &Orphamine&3,210&5,639&722,975\\
     &Covtype&581,012&54&1,277,532\\
     &US\_Census&2,458,285&68&6,684,779\\
     &NYT-2000&185,771&2,000&14,863,292\\
\hline
&Foreign\_Exchange&5,015&22&4,431\\
&Parkinsons&5,875&20&4,655\\
&Google\_Review&5,456&24&5,110\\
&Ozone&2,534&71&6,642\\
&Drive\_Signal&58,509&48&112,403\\
&Health\_Examination&9,813&46&23,678\\
\textbf{Continuous}&Health\_Lab&9,813&168&32,814\\
\&&Musk&6,598&166&43,792\\
\textbf{discretized}&Motion\_Capture&78,096&30&71,394\\
&Superconductor&21,263&82&69,800\\
&Metabolic\_Net&65,5543&23&60,058\\
&Bankruptcy&43,405&64&109,745\\
&Protein&9,672&806&311,619\\
&WECs&287,999&49&565,421\\
&Facebook&603,813&33&796,875\\
    \bottomrule
  \end{tabular}
    \caption{Summary of the datasets used in the experiment with respect to the data type, number of data instances, number of attributes and number of target values in the test datasets when the masking rate is 20\% for these datasets.}\label{tab:data_summary}
\end{table}


\section{Experiments - Comparison Study}

In this section, we compare and analyze the performance of a range of graphical and self-supervised learning models with respect to target-agnostic prediction for three types of attributes, which are categorical, continuous and discretized, over in total 30 medium to large-scale datasets. Each of these datasets possesses only one of the three types of attributes. These datasets are provided by the UCI repository of machine learning \citep{Dua:2019}, KEEL data repository \citep{alcala2011keel} and the Kaggle data repository\footnote{https://www.kaggle.com/datasets}. Table~\ref{tab:data_summary} summarizes the main characteristics of these datasets. Apart from the target-agnostic prediction performance, we also evaluate the scalability of each model against the size of the attribute set and the size of the data in terms of the amounts of training and prediction time.

\subsection{Experimental Setting}\label{sec:experiment_settings}
There are seven target-agnostic learning approaches competing throughout the experiments, which are the Chordalysis, VAEAC, GAIN, MIDAS, MADE, BERT and XLNet. We also include a naive baseline method for each type of data, which verifies whether the performance of each competing approach is meaningful. For the categorical and discretized data, it is the most frequent value of each attribute, referred to as the \textbf{Most\_Freq}, while for the continuous data, it is the median value of each attribute, referred to as the \textbf{Median}. 

Except for the generative graphical model Chordalysis, the training of the rest of the target-agnostic models is performed with 20\% masking rate. This means that the values of 20\% of the attributes (with rounding up) of each training instance are randomly chosen to be replaced by zero. The masking is performed regardless of the missing values in the data. The models are then trained to recover the values of these masked attributes (for each training instance) with 25\% of them (with rounding up) being used as the validation target attributes\footnote{The number of validation target attributes for the dataset with the least number of attributes: Chess, is 1 as we set its training target attributes to be 2 for every instance.} for optimizing the settings of each model.

As for the settings of each model for the experiments, Chordalysis has the simplest ones. For the scoring methods, we have adopted both the SMT and the QNML methods to run the experiments and choose the better results to report for each dataset. A missing value is treated as a separate attribute value and taken into
account by Chordalysis for inference exactly like other values. For the original implementation of Chordalysis, each target value has to be predicted sequentially for each test data instance, which turns out to be slow for large-scale datasets. Therefore, to accelerate the prediction, we have also implemented parallel prediction for Chordalysis by dividing the test dataset into equal-sized chunks and running the prediction on each chunk in parallel with multiple processes. We observed that this implementation speeds up the prediction time of Chordalysis by an order of magnitude.  

Unless mentioned otherwise, we adopt the settings specified in Table~\ref{tab:hyperparams} for each deep learning model that are necessary to understand and reproduce their results. More specifically, the table summarizes the common hyper-parameter value ranges used for the hyper-parameter optimization for these models over the validation target attributes of the training data. Note that for BERT and XLNet, the number of layers correspond to the number of transformers in them, while the number of neurons is specific to the feed-forward neural network within each transformer. Apart from these common hyper-parameters, some of the models also possess their unique sets of hyper-parameters to be optimized. 

More specifically, 
for the GAIN model, the hint rate, which corresponds to the percentage of masked attributes revealed to the discriminator, is chosen from the set \{0.1,0.2,...,0.9\}; the trade-off hyper-parameter attached to the reconstruction loss of the generator on the observed attributes is chosen from \{0.01, 0.1, 1,...,$10^4$\}. For the MIDAS model, the dropout ratio for each hidden layer is chosen from \{0.1,0.2,...,0.5\}. For the BERT model, the number of heads for multi-head self-attention is chosen from \{16,32,...,256\}. For XLNet, the same hyper-parameters are chosen from the same ranges as in BERT; the length of memory is chosen from \{32,64,...,256\}. For both BERT and XLNet, the number of hidden layers for each of their feed-forward neural networks within a transformer is set to be 1.

When dealing with the categorical/discretized data, except for BERT and XLNet which take in integers that represent each category value, and Chordalysis which takes in the category values directly, the other models take in the binarized versions of the category values given their implementations. For example, if an attribute takes on the set of values \{A, B, C\}, then it will be converted into three corresponding binary attributes and its resulting binarized values will be fed into those models. In this case, for a categorical attribute, a dedicated softmax computation is performed over the part of the network outputs corresponding to its binary attributes for predicting its value. 

Finally, either the official implementations or the credible re-implementations\footnote{They are used when the original implementations are not available or unnecessarily complicated for our experiments.} of the competing models have been used for the experiments. The entire experiments have been conducted on a Dell PowerEdge R940 server with a 64-core Intel(R) Xeon(R) Gold 6145 CPU @ 3.00GHZ and 512 GiB of memory.

\begin{table}[t]
    \centering
    \begin{tabular}{lccccc}
    \toprule
        \textbf{Models} & \#\textbf{Latent Dim.} &
        \#\textbf{Layers} & \#\textbf{Neurons Per Layer} &\textbf{Batch Size}\\
        \hline
        VAEAC&\{32,64,128,...,512\}&1~to~4&\{32,64,128,...,512\}&\{8,16,...,64\}\\
        GAIN&N/A&1~to~3&\{\#Attributes,32,64,...,512\}&\{8,16,...,128\}\\
        MADE&N/A&1~to~3&\{32,64,128,...,2048\}&\{8,16,...,128\}\\
        MIDAS&N/A&1~to~3&\{32,64,128,...,512\}&\{8,16,...,64\}&\\
        BERT&\{32,64,...,256\}&1~to~12&\{32,64,...,256\}&\{8,16,...,64\}\\
        XLNet&\{32,64,...,256\}&1~to~12&\{32,64,...,256\}&\{8,16,...,64\}\\

    \bottomrule
    \end{tabular}
    \caption{The common hyper-parameter value ranges of each competing deep learning models for the hyper-parameter optimization on the validation datasets.}
    \label{tab:hyperparams}
\end{table}

\subsection{Target Agnostic Prediction}
We compare the different models on two evaluation metrics: the weighted average percentage of misclassification (WAPMC) for categorical and discretized datasets, and the weighted normalized root mean squred error (WNRMSE) for continuous datasets. More specifically, the definition of the WAPMC metric is as follows:  

\begin{equation}
\begin{split}
    \textbf{WAPMC}:  &\sum_{j\,:\,X_j\in\boldsymbol{V}}\frac{M_{j}}{M}{\sum^{n}_{i=1}\Bigg(\mathbbm{1}\Big(X^{(i)}_{j}\neq``?" \wedge \widetilde{X}^{(i)}_{j}\in\widetilde{\boldsymbol{X}}^{(i)}_{m}\wedge X^{(i)}_{j}\neq\hat{X}^{(i)}_{j}\Big)\Bigg)};\\
    &M_j = {\sum^{n}_{i=1}\Bigg(\mathbbm{1}\Big(X^{(i)}_{j}\neq``?" \wedge \widetilde{X}^{(i)}_{j}\in\widetilde{\boldsymbol{X}}^{(i)}_{m}\Big)\Bigg)};\\ &M=\sum_{k\,:\,X_k\in\boldsymbol{V}}M_k
\end{split}
\end{equation}
where $\mathbbm{1}(\cdot)$ is the indicator function; the symbol ``?" represents the missing value; $M_j$ is the number of data instances that are originally observed at the $j$-th feature and then masked at this feature for the testing; $M$ is the total number of target values to be predicted for the test data; $\frac{M_j}{M}$ is thus the weight for the $j$-th attribute which will be different for each attribute when missing values exist in them; for the $i$-th instance, $\hat{X}^{(i)}_{j}$ stands for the predicted value for the target attribute $X_{j}$'s original value $X^{(i)}_{j}$. Similarly, the metric WNRMSE for continuous data is defined as follows:
\begin{equation}
\begin{split}
    &\textbf{WNRMSE}:\\  &\sum_{j:X_j\in\boldsymbol{V}}\frac{M_{j}}{M}\sqrt{\frac{1}{M_j}{\sum^{n}_{i=1}\Bigg(\mathbbm{1}\Big(X^{(i)}_{j}\neq``?" \wedge \widetilde{X}^{(i)}_{j}\in\widetilde{\boldsymbol{X}}^{(i)}_{m}\Big)\times\Big(X^{(i)}_{j}-\hat{X}^{(i)}_{j}\Big)^2\Bigg)}}\Bigg/\sigma(X_{j}) 
\end{split}
\end{equation}
Here, the normalization of weighted RMSE is taken for each attribute $X_j$ by dividing their standard deviations $\sigma(X_j)$ calculated from the training data. In this case, the larger variance the values of an attribute has, the less impact the goodness of its predictions will have on the evaluation of the model performance. As a result of this normalization step, predictions on each attribute will tend to have similar degrees of impact on the performance evaluation.

For comprehensive comparisons of the different models on target-agnostic prediction, we test the robustness of their performance with a range of masking rates for the test data which are 10\%, 20\%, 40\%, 60\% and 80\%. More specifically, we first train each model with the 20\% masking rate on the input attributes of the training data as specified in Section~\ref{sec:experiment_settings} and with hyper-parameter optimization on the candidate settings specified in Table~\ref{tab:hyperparams}. Once the training process is finished, the test data is then masked with each of the above percentages and fed into each model for predicting the values of the masked attributes. Except for Chordalysis, all the models have adopted multiple imputation for predicting the target attributes in the test data. We set the number of imputation samples to be 10, which means that the final prediction for each target attribute is averaged over 10 imputed sample values generated by each model. Finally, we adopt the 80\%-20\% split ratio to create the training and test data for each dataset in the experiments.

\begin{figure}[t]
\centering  
\includegraphics[width=6in]{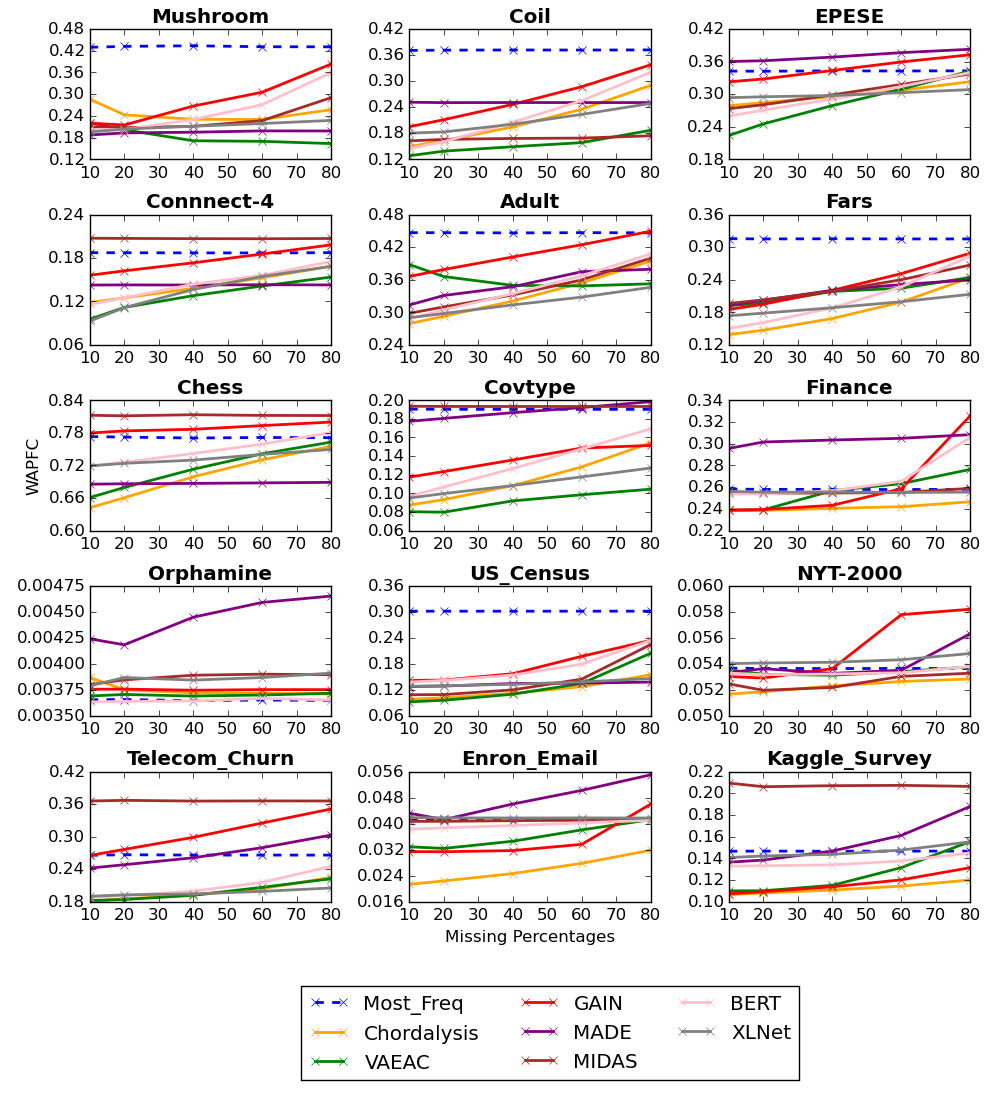}
\caption{The WAPMC results of each model across the 15 \textbf{categorical} datasets with different percentages of masking for their test datasets.}\label{fig:wapmc_results}
\end{figure}

\subsection{Results on Categorical Data}
Figure \ref{fig:wapmc_results} displays the WAPMC results of each model across the different masking rates for the test data of the 15 categorical datasets. It shows that overall, Chordalysis and VAEAC are the two best-performing models, although they tend to become less robust when the masking rate on the test data increases. 

\begin{figure}[!htbp]
    \begin{subfigure}[t]{0.95\textwidth}
        \centering
        \includegraphics[height=2.2in]{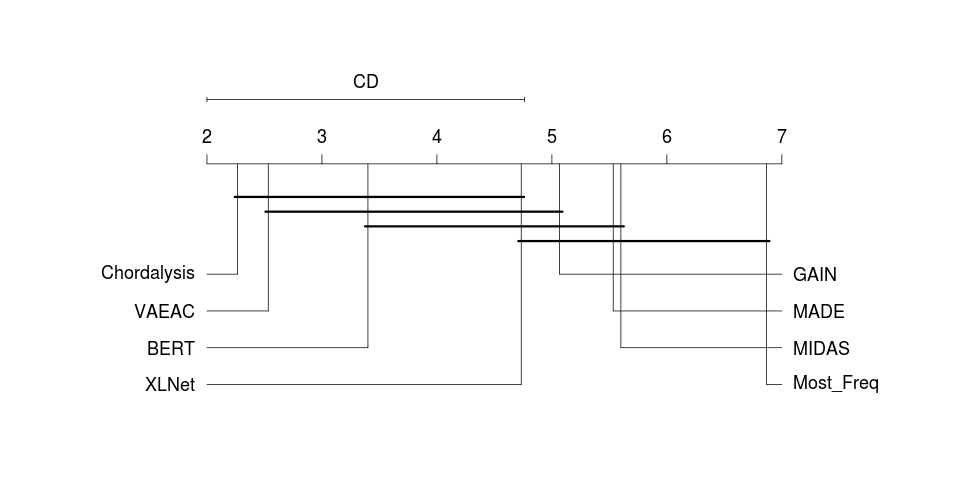}
        \vspace*{-25pt}\caption{The CD diagram for 20\% masking rate}
    \end{subfigure}%
    \\
            \begin{subfigure}[t]{0.95\textwidth}
        \centering
        \includegraphics[height=2.2in]{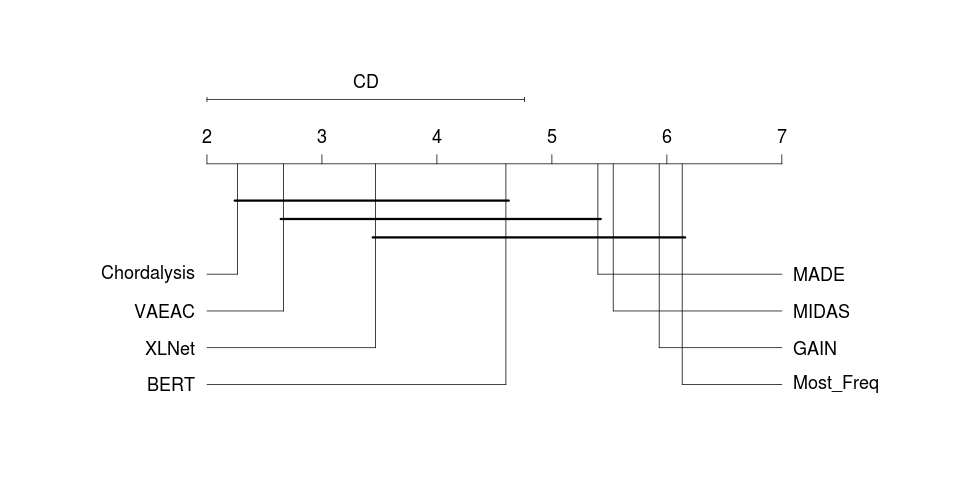}
        \vspace*{-25pt}\caption{The CD diagram for 60\% masking rate}
    \end{subfigure}%
    \\
            \begin{subfigure}[t]{0.95\textwidth}
        \centering
        \includegraphics[height=2.2in]{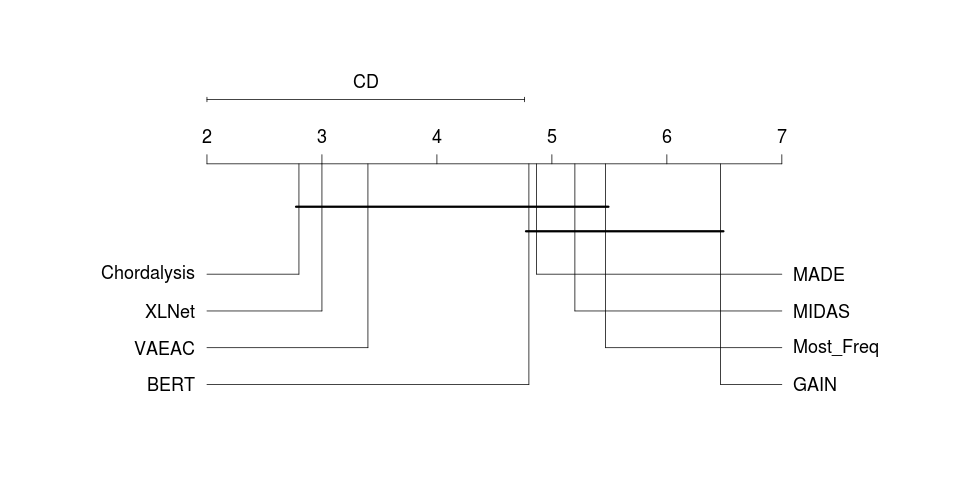}
        \vspace*{-25pt}\caption{The CD diagram for 80\% masking rate}
    \end{subfigure}%
    \caption{The critical difference (CD) diagrams show the mean ranks of each model under the different masking rates for the test data of the 15 \textbf{categorical} datasets. The lower the rank (further to the left) the better performance of a model under the particular masking rate compared to the others on average. A line in each diagram indicates that there is no significant difference in performance among the models crossed by that particular line in terms of the Friedman test that compares the ranks of multiple classifiers~\citep{demvsar2006statistical}.}\label{fig:cd_digrams_cat}
\end{figure}

The above finding is confirmed by the CD diagram results in Figure~\ref{fig:cd_digrams_cat}. Chordalysis is statistically different from four other models (i.e. GAIN, MADE, MIDAS and Most\_Freq) in performance at the 20\% and 60\% masking rates, and from one model (i.e. GAIN) at the 80\%. For VAEAC, it is statistically different from three other models at the 20\% and 60\%, and from GAIN at the 80\%. Furthermore, VAEAC is less robust than Chordalysis as its mean rank has decreased from 2.53 at the 20\% masking rate to 3.4 at the 80\%, while Chordalysis has experienced a relatively minor decrease from 2.27 to 2.8. The reason for VAEAC being less robust might be because its training with the 20\% masking rate might not have been well regularized as its KL-divergence term is not scaled in this case to properly penalize its loss function. This can result in VAEAC not generalizing well to recover larger sets of masked attributes based on a much smaller set of observed attributes in the test datasets.

The above observations can also be found on the BERT model. As shown by Figure~\ref{fig:cd_digrams_cat}, it tends to perform better on lower masking rates but worse on larger ones. In comparison, the performance of XLNet appears to be more robust on predicting a larger set of target attributes compared to VAEAC and BERT. This is supported by the fact that the mean rank of XLNet has increased from 4.73 at the 20\% masking rate to 3.0 at the 80\%, which surpasses the mean ranks of VAEAC and BERT (i.e. 3.4 and 4.8 respectively at the 80\%). 

The above findings suggest that the independence assumption BERT imposes on its prediction might become too simple to deal with larger sets of masked attributes. In contrast, the permutation-based autoregressive modelling of XLNet is likely to have exploited the correlations between even a small set of observed attributes to maintain its predictive performance. This is further evidenced by the performance of XLNet on datasets with small numbers of attributes (i.e. see the performance on Chess, Adult, Telecom\_Churn and Fars datasets in Figure~\ref{fig:wapmc_results}) under the 80\% masking rate. In this case, XLNet's performance turns out to be better than the performance of Chordalysis, VAEAC and BERT.

Similar results can be found on the MADE model. First, the mean rank of MADE has increased from 5.53 at the 20\% masking rate to 4.86 at the 80\%. Furthermore, its performance on datasets with small numbers of attributes under the 60\% and 80\% masking rates is overall comparable to the performance of the previous models. These results are somewhat expected as both MADE and XLNet leverage the permutation-based autoregressive modelling for their prediction. The superiority of XLNet in performance in this case might be attributed to its stacked transformer-XL architecture. 

The performance of the MIDAS model is overall the most stable (as shown by its flat performance lines in Figure~\ref{fig:wapmc_results}) but in general inferior to the performance of the previous models. Such stability might be due to the dropout mechanism of MIDAS applied to its hidden layers which regularizes its performance. For this experiment, the least robust model is GAIN whose mean rank is at the 5th place for the 20\% masking rate but declines to the bottom for the 80\%. This finding coincides with the observation from \citet{ivanov2018variational} in their experiments; when the training data is highly observed (corresponding to the 20\% masking rate in our experiments) and the testing data is highly unobserved (i.e. at the 80\% masking rate), GAIN turns into a disadvantage. This might be because GAIN has not been designed to learn to recover the ground-truth values of the masked attributes during the training, which makes it fail to deal with the highly unobserved test data.

\begin{figure}[!htbp]
\centering
    \begin{subfigure}[t]{0.4\textwidth}
        \centering
        \includegraphics[height=2.5in]{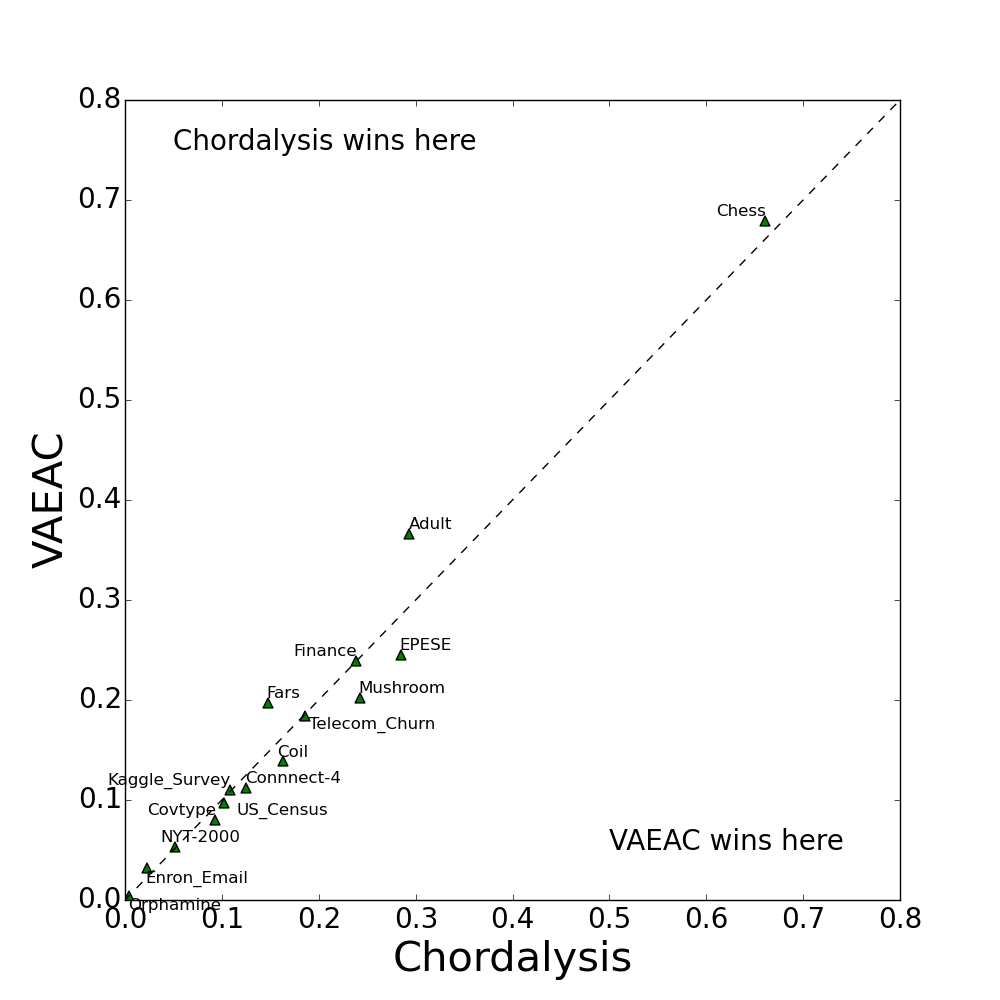}
    \end{subfigure}%
        \begin{subfigure}[t]{0.4\textwidth}
        \centering
        \includegraphics[height=2.5in]{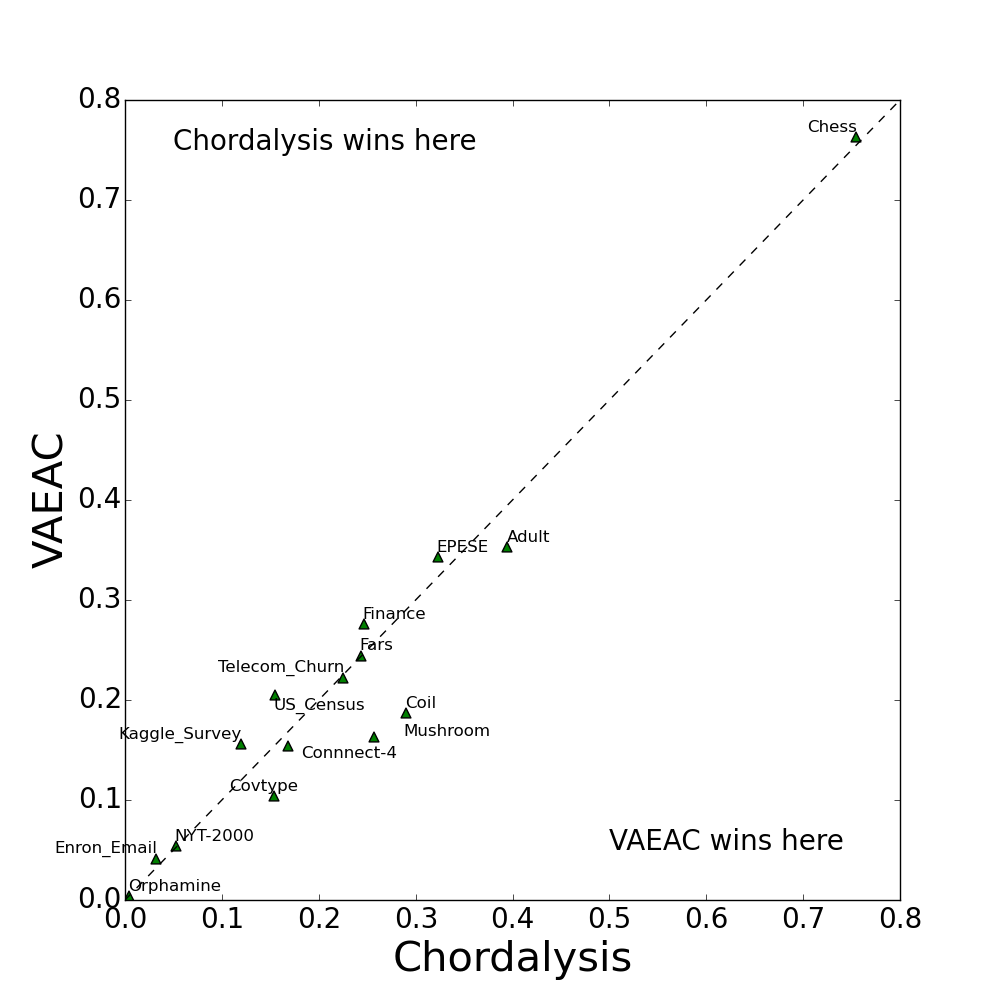}
    \end{subfigure}%
    \\
        \begin{subfigure}[t]{0.4\textwidth}
        \centering
        \includegraphics[height=2.5in]{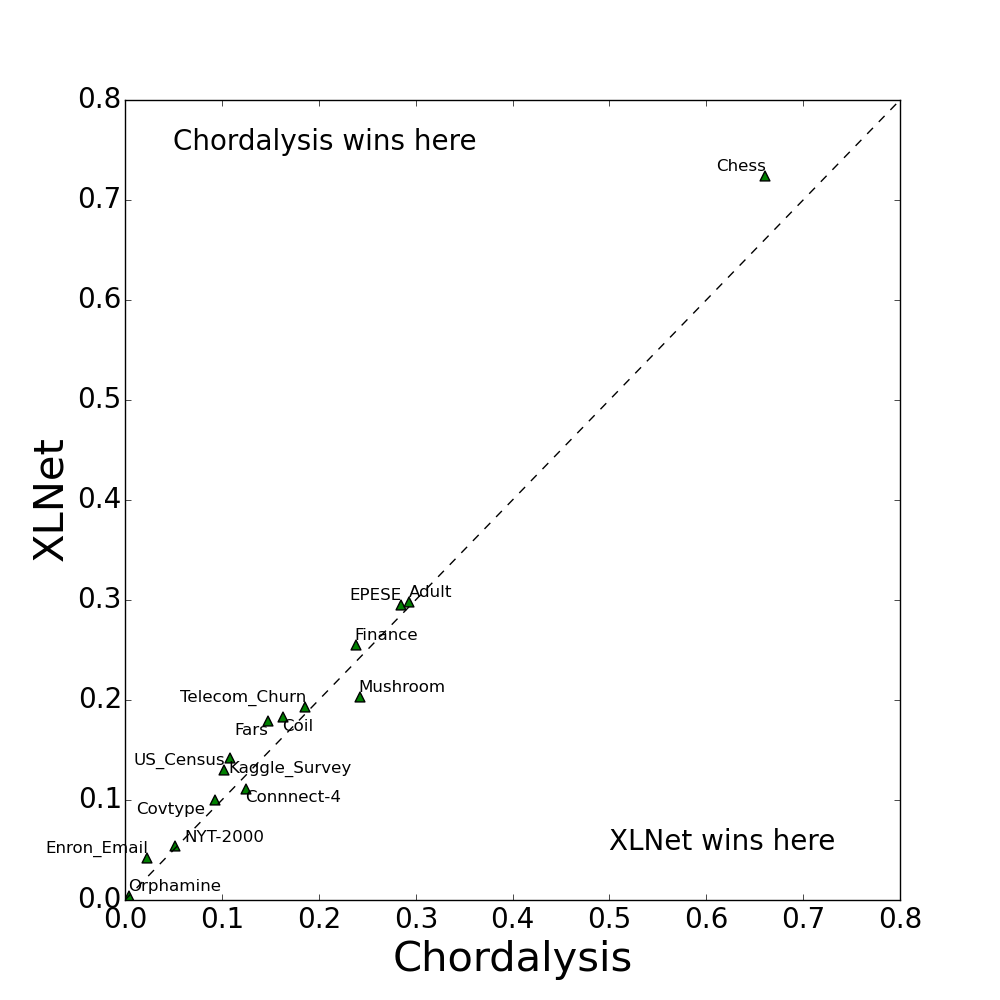}
    \end{subfigure}%
        \begin{subfigure}[t]{0.4\textwidth}
        \centering
        \includegraphics[height=2.5in]{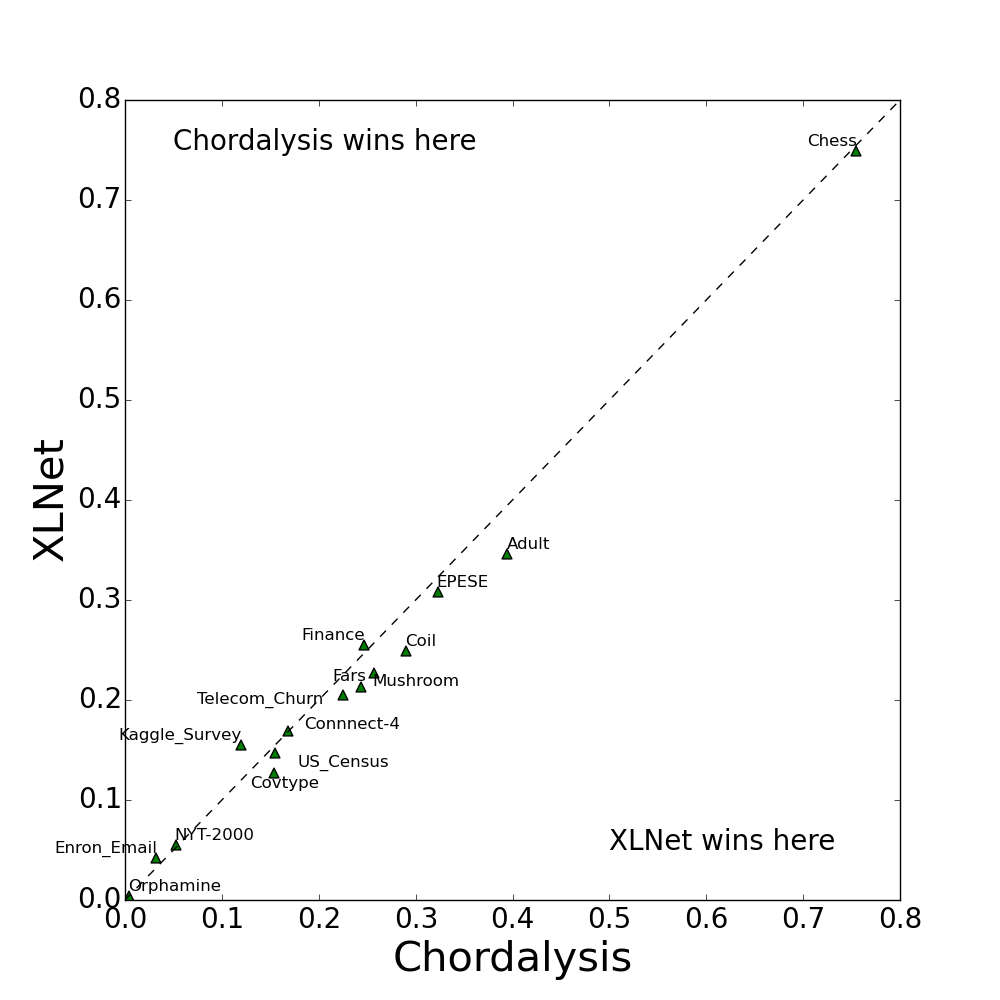}
    \end{subfigure}%
    \\
    \begin{subfigure}[t]{0.4\textwidth}
        \centering
        \includegraphics[height=2.5in]{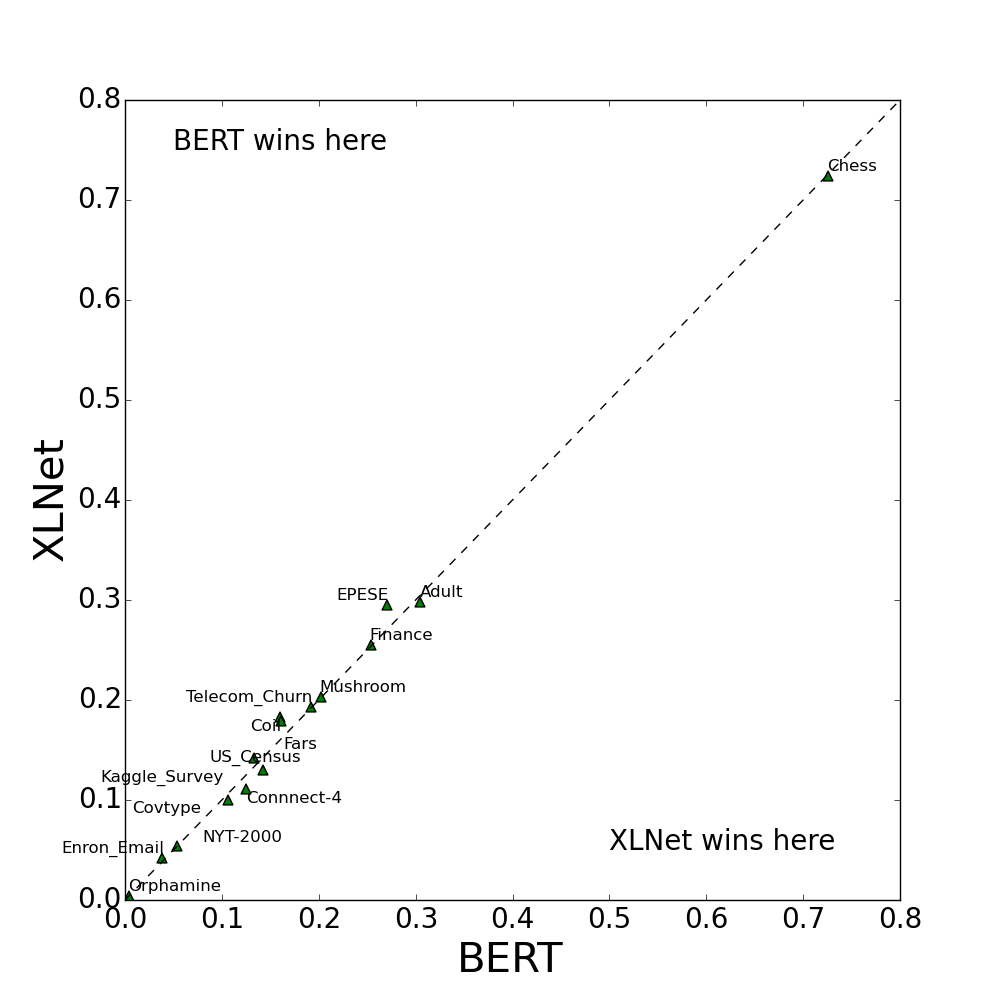}
        \caption{20\% masking rate}
    \end{subfigure}%
        \begin{subfigure}[t]{0.4\textwidth}
        \centering
        \includegraphics[height=2.5in]{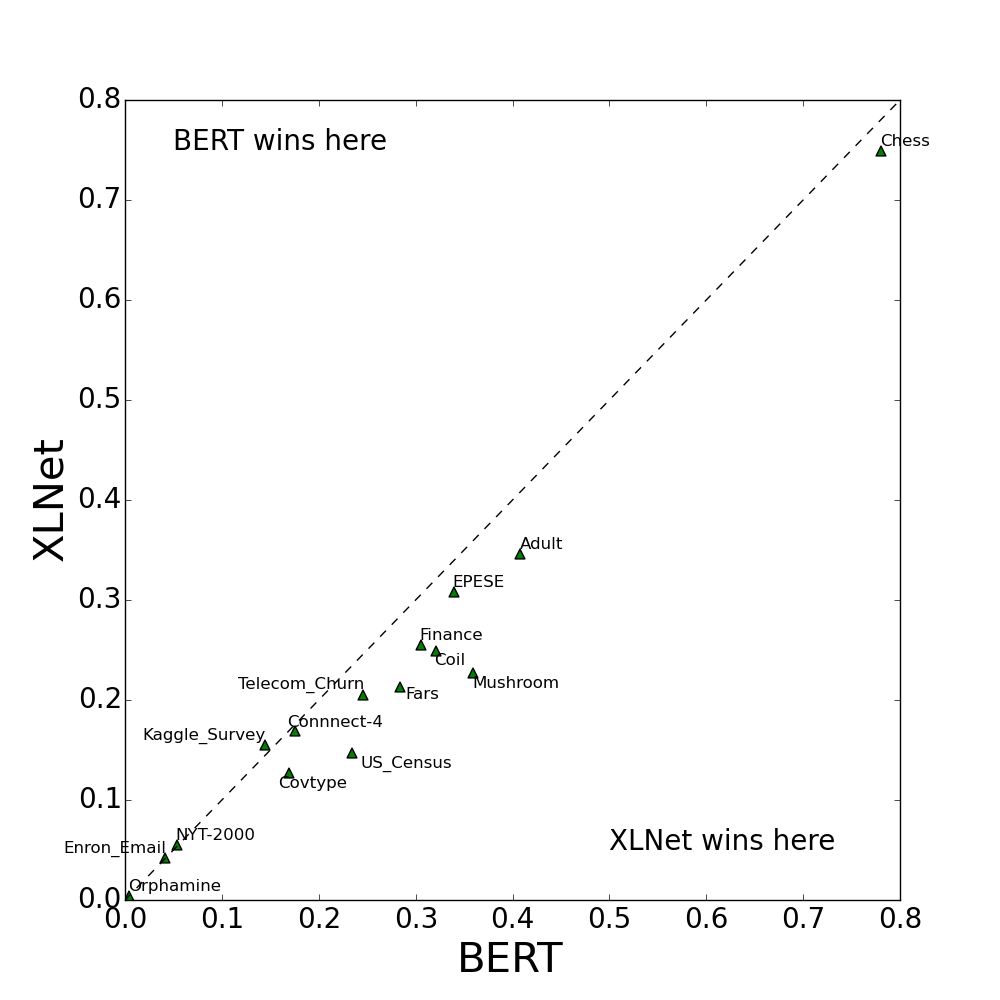}
       \caption{80\% masking rate}
    \end{subfigure}%
    \caption{Comparison of WAPMC for Chordalysis \textit{versus} VAEAC, Chordalysis \textit{versus} XLNet and BERT \textit{versus} XLNet on the 15 \textbf{categorical} datasets.}\label{fig:wapmc_scatter_plot}
\end{figure}

To further compare the WAPMC results of the best-performing models, Figure \ref{fig:wapmc_scatter_plot} presents the scatter plots of the pairwise WAPMC of different pairs of models over the 20\% and 80\% masking rates. Each point in the plots represent a dataset. For Chordalysis versus VAEAC, data points move further away from the diagonal line (corresponding to tied performance between the two models) when the masking rate for the test data increases from 20\% to 80\%. There is also a moderate trend that Chordalysis wins more often, i.e. 4 at the 20\% against 6 at the 80\%, although the wins on the Enron\_Email and Chess are marginal. We can also see that even though Chordalysis has higher mean ranks at both masking rates according to Figure \ref{fig:cd_digrams_cat}, it has fewer to equal numbers of wins compared to VAEAC. 

For Chordalysis versus XLNet, we can observe that the former wins more under the 20\% masking rate, while the latter wins more under the 80\%. However, the margins of the wins are not significant in both cases. In comparison, the (absolute) difference in the mean ranks between the two, as shown by Figure \ref{fig:cd_digrams_cat}, has reduced significantly from 2.47 at the 20\% to 0.2 at the 80\%. For BERT versus XLNet, the two appear to be close in performance in the first place, and then XLNet dominates the number of wins under the 80\% masking rate. The margins of the wins are significant in this case for XLNet over BERT. This observation is consistent with the difference in the mean ranks between the two; XLNet is 1.8 higher than BERT at the 80\%. However, under the 20\% masking rate, this consistency no longer exists as BERT is 1.33 higher than XLNet in the mean rank.

\begin{figure}[!htbp]
\centering  
\includegraphics[width=6in]{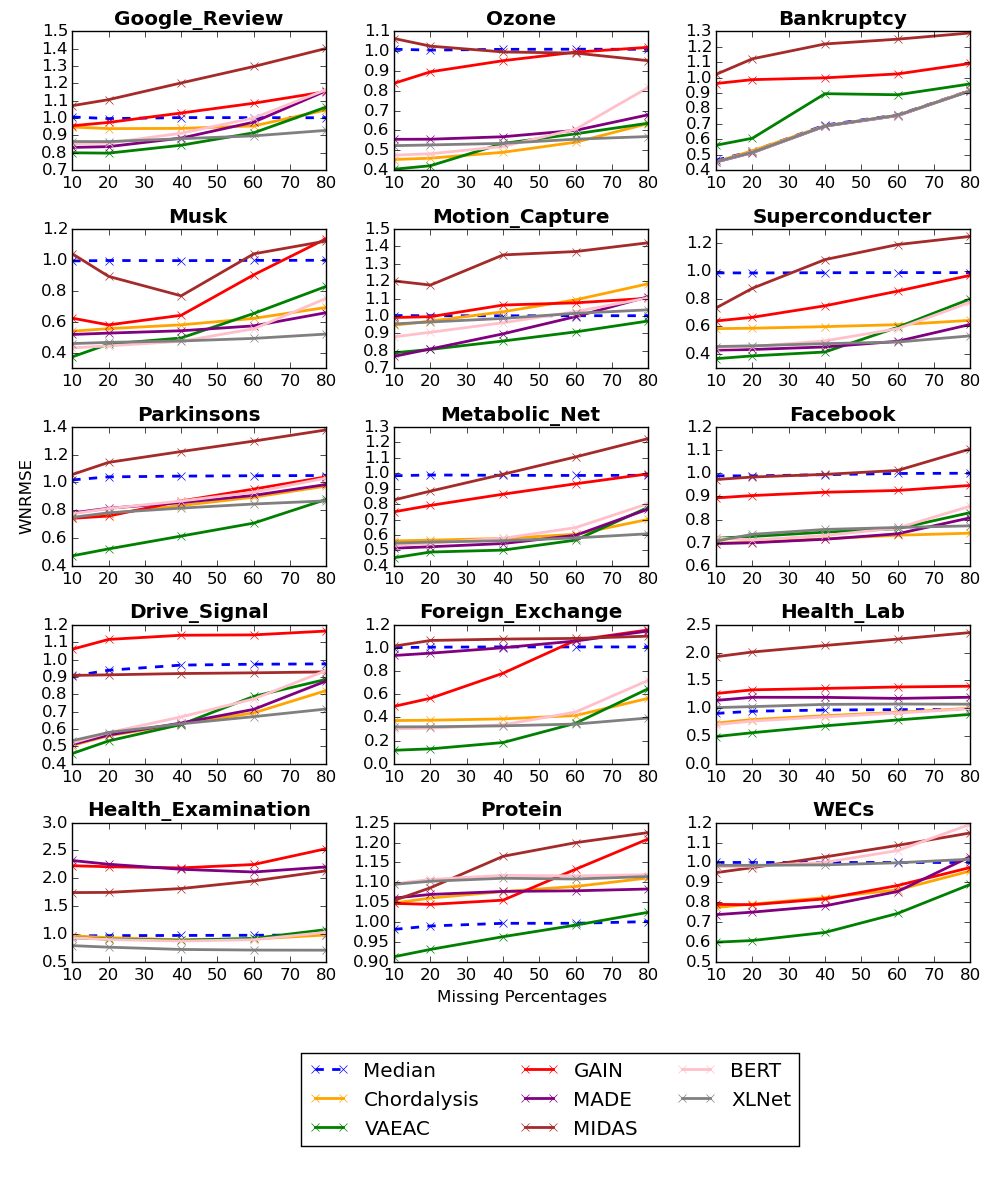}
\caption{The WNRMSE results of each model across the 15 \textbf{continuous} datasets under the different masking rates for their test datasets.}\label{fig:wnrmse_resuts}
\end{figure}

\subsection{Results on Continuous Data}
\begin{figure}[!htbp]
    \begin{subfigure}[t]{0.95\textwidth}
        \centering
        \includegraphics[height=2.2in]{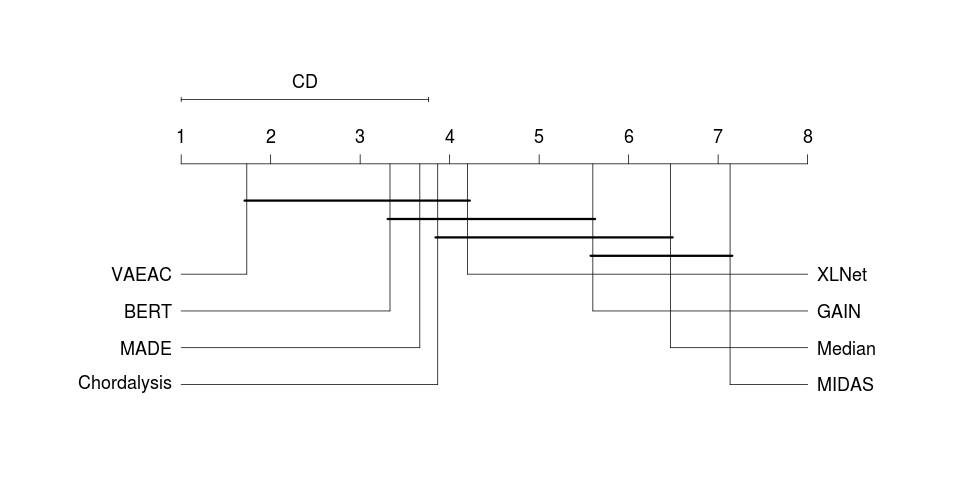}
        \vspace*{-25pt}\caption{The CD diagram for 20\% masking rate}
    \end{subfigure}%
    \\
            \begin{subfigure}[t]{0.95\textwidth}
        \centering
        \includegraphics[height=2.2in]{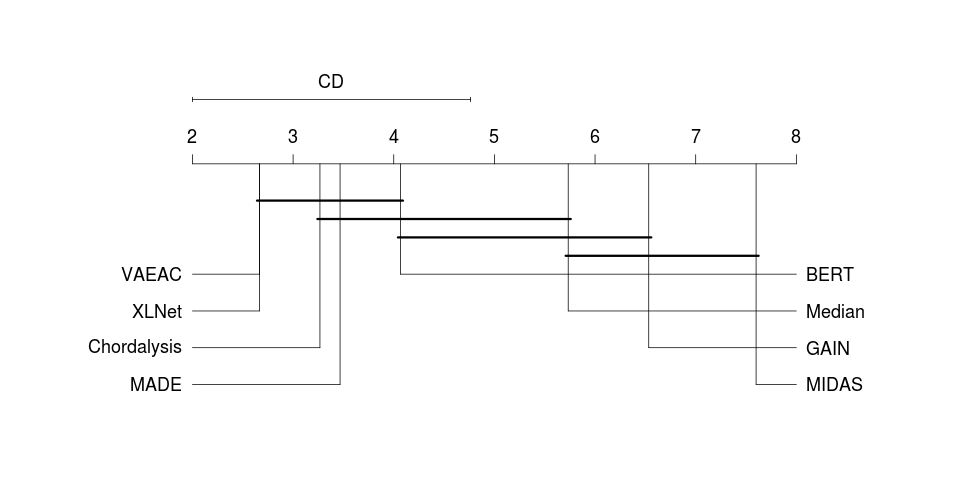}
        \vspace*{-25pt}\caption{The CD diagram for 60\% masking rate}
    \end{subfigure}%
    \\
            \begin{subfigure}[t]{0.95\textwidth}
        \centering
        \includegraphics[height=2.2in]{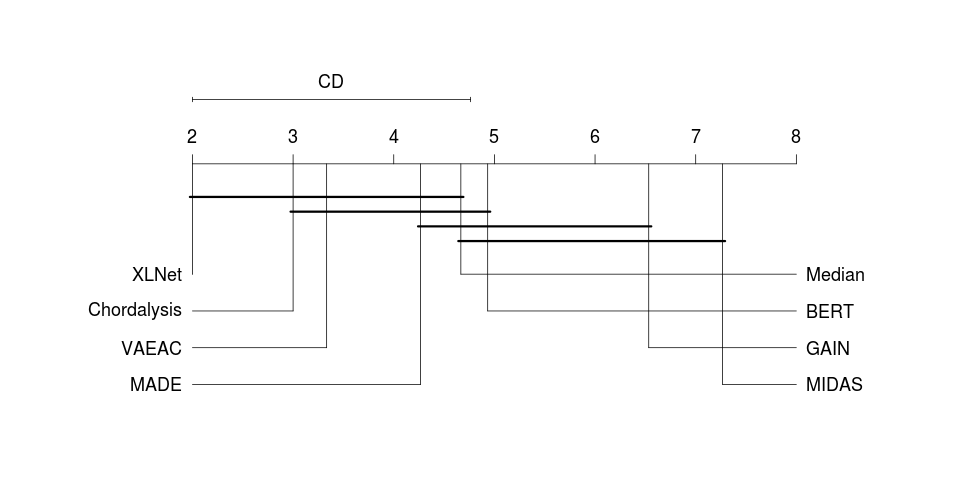}
        \vspace*{-25pt}\caption{The CD diagram for 80\% masking rate}
    \end{subfigure}%
    \caption{The critical difference (CD) diagrams show the average ranks of each model under the different masking rates across the 15 \textbf{continuous} datasets.}\label{fig:cd_digrams_cont}
\end{figure}

Figure \ref{fig:wnrmse_resuts} shows the WNRMSE results of each model under the different masking rates for the test data of the 15 continuous datasets. We can see that overall, VAEAC is the best model when the masking rates are low, while XLNet takes over the other models when the masking rates are high. This is further supported by the CD diagram results in Figure \ref{fig:cd_digrams_cont}. On average, VAEAC is the top-ranked model at the 20\% masking rate and its mean rank is 1.6 higher than that of the second-best model which is BERT. However, it becomes the third-best model under the 80\% while XLNet comes to the first place in terms of the mean rank. In between at the 60\% masking rate, VAEAC and XLNet tie in the mean rank at 2.67. 

These findings correspond to the previous observations for VAEAC and XLNet on the categorical datasets. Therefore, they can be explained in the same ways; VAEAC might not generalize well to highly masked data due to its unscaled KL-divergence that might fail to perform proper regularization during the training on highly observed data; XLNet leverages the permutation-based autoregressive modelling that can exploit more correlations between the attributes, which helps it perform robustly on the highly masked test data. Another similarity we can find in the results between the categorical and continuous data is that BERT has the same trend in performance as VAEAC throughout the different masking rates; It performs better on the lower masking rates but worse on the higher rates, again most likely due to its independence assumption on the prediction.

\begin{figure}[!htbp]
\centering
    \begin{subfigure}[t]{0.4\textwidth}
        \centering
        \includegraphics[height=2.5in]{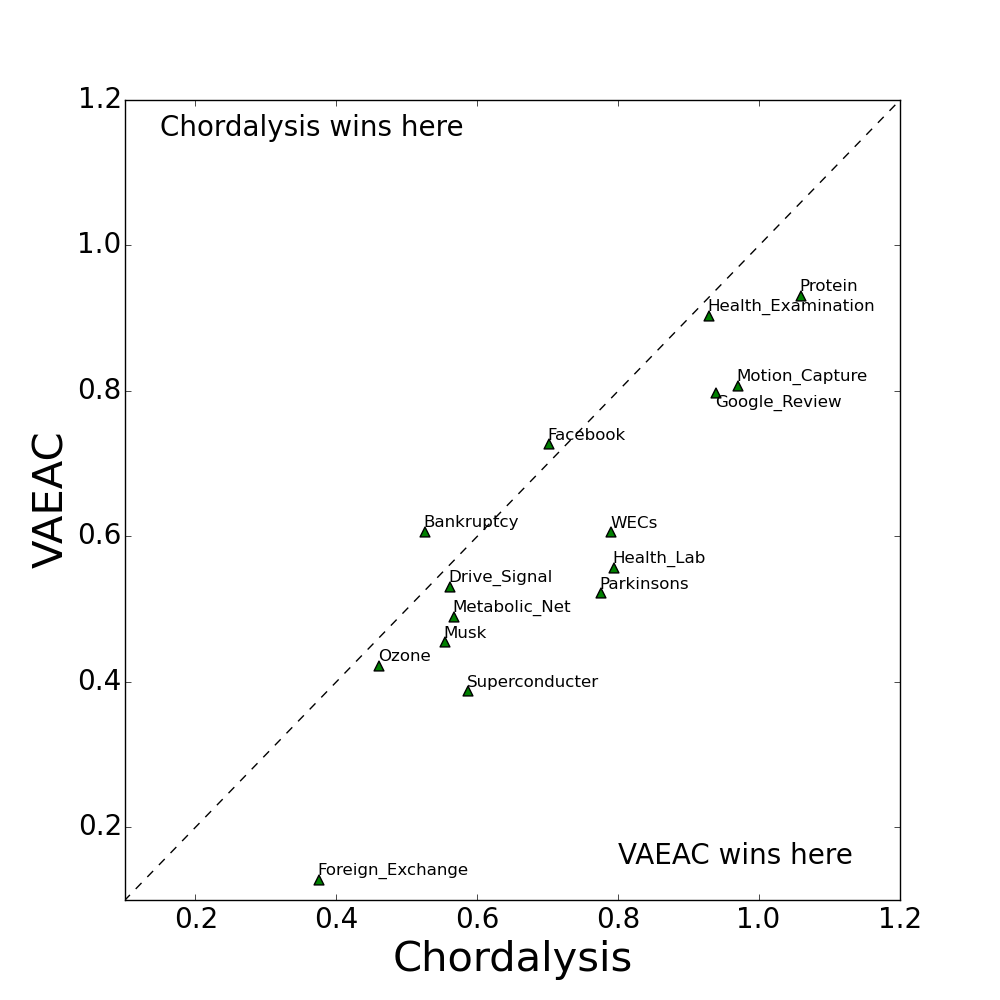}
    \end{subfigure}%
        \begin{subfigure}[t]{0.4\textwidth}
        \centering
        \includegraphics[height=2.5in]{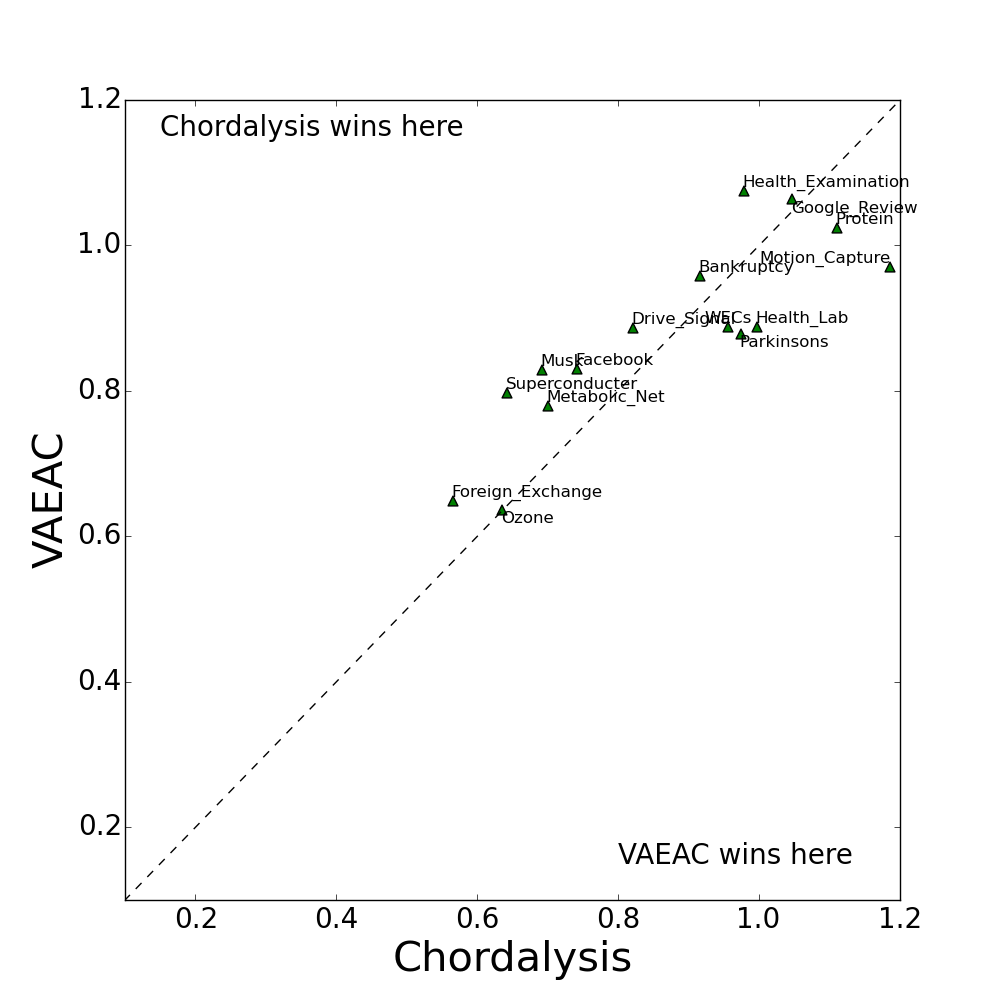}
    \end{subfigure}%
    \\
        \begin{subfigure}[t]{0.4\textwidth}
        \centering
        \includegraphics[height=2.5in]{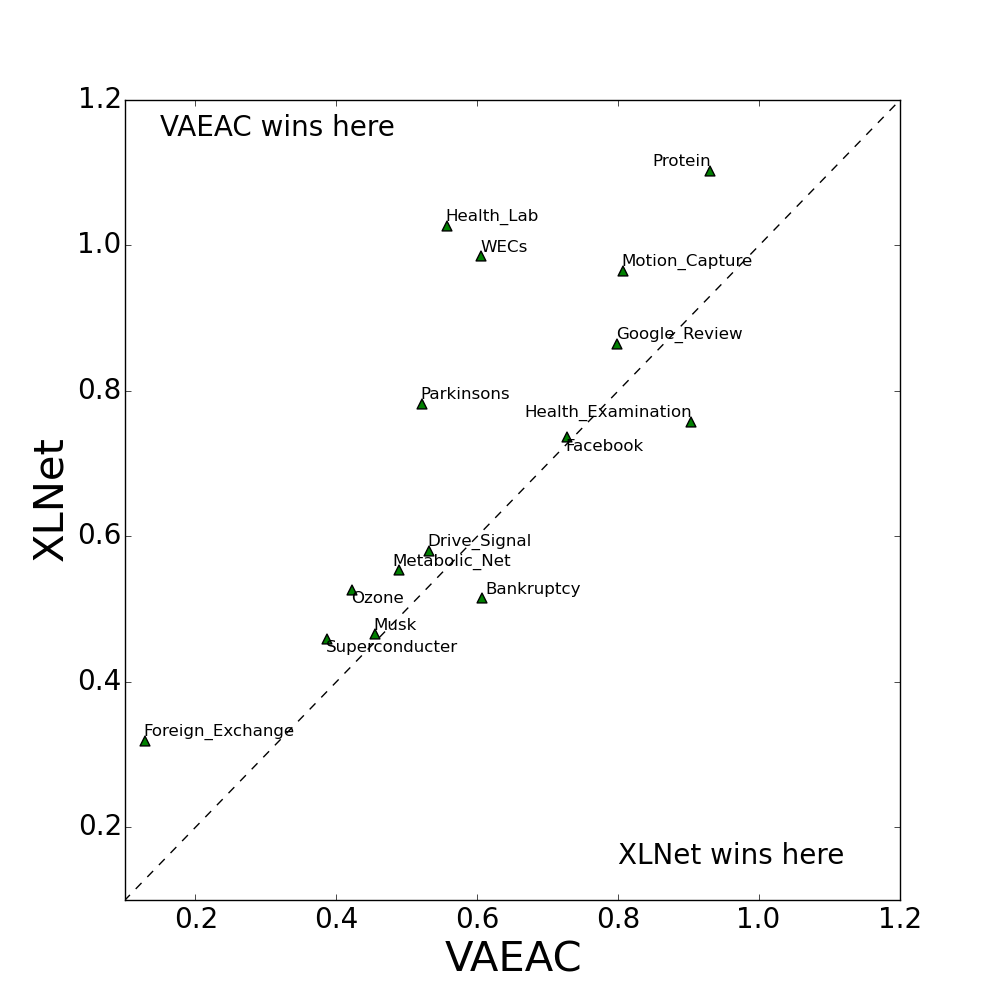}
    \end{subfigure}%
        \begin{subfigure}[t]{0.4\textwidth}
        \centering
        \includegraphics[height=2.5in]{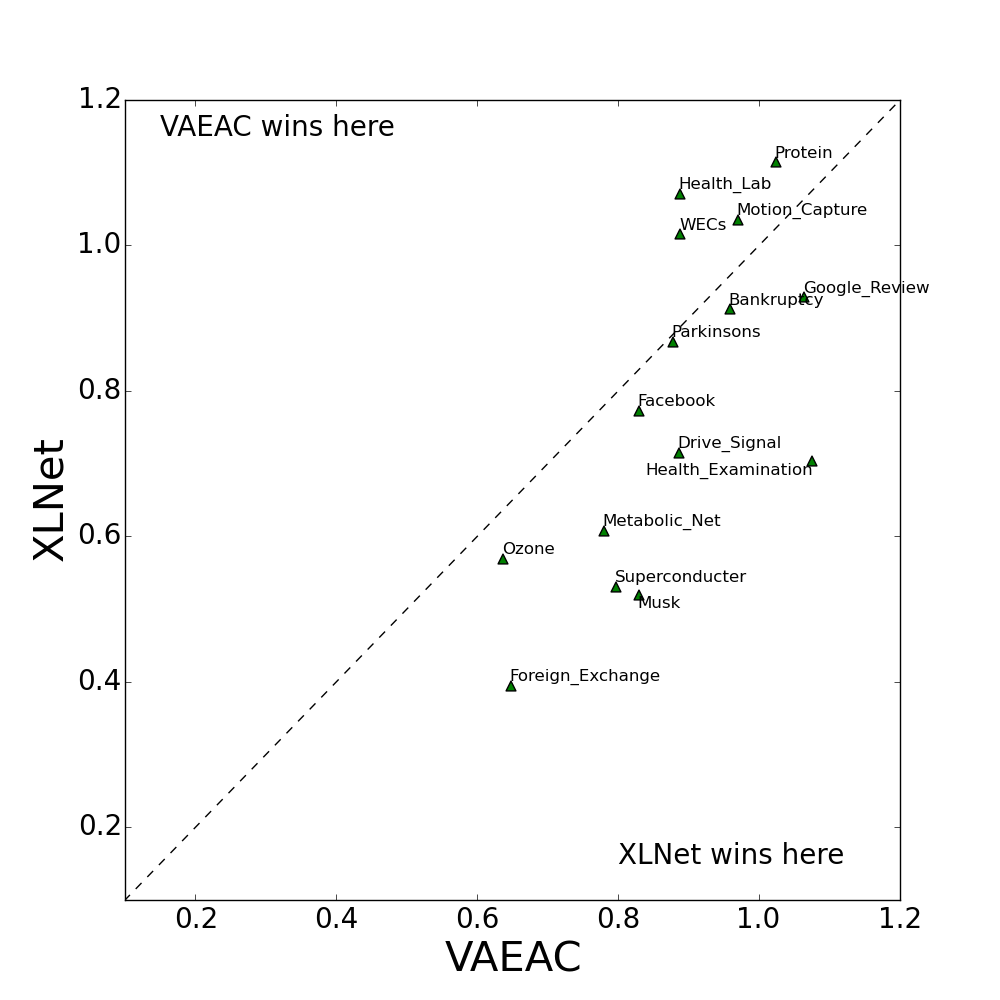}
    \end{subfigure}%
    \\
        \begin{subfigure}[t]{0.4\textwidth}
        \centering
        \includegraphics[height=2.5in]{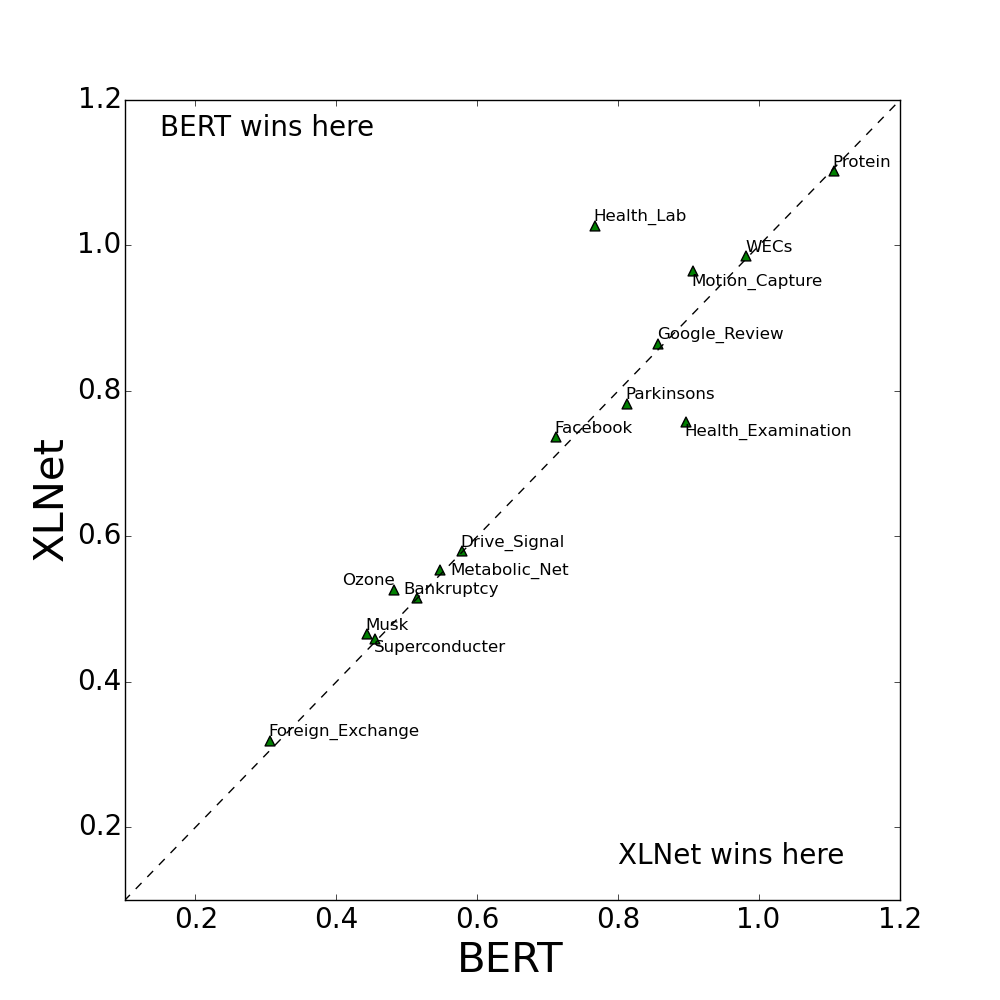}
    \end{subfigure}%
        \begin{subfigure}[t]{0.4\textwidth}
        \centering
        \includegraphics[height=2.5in]{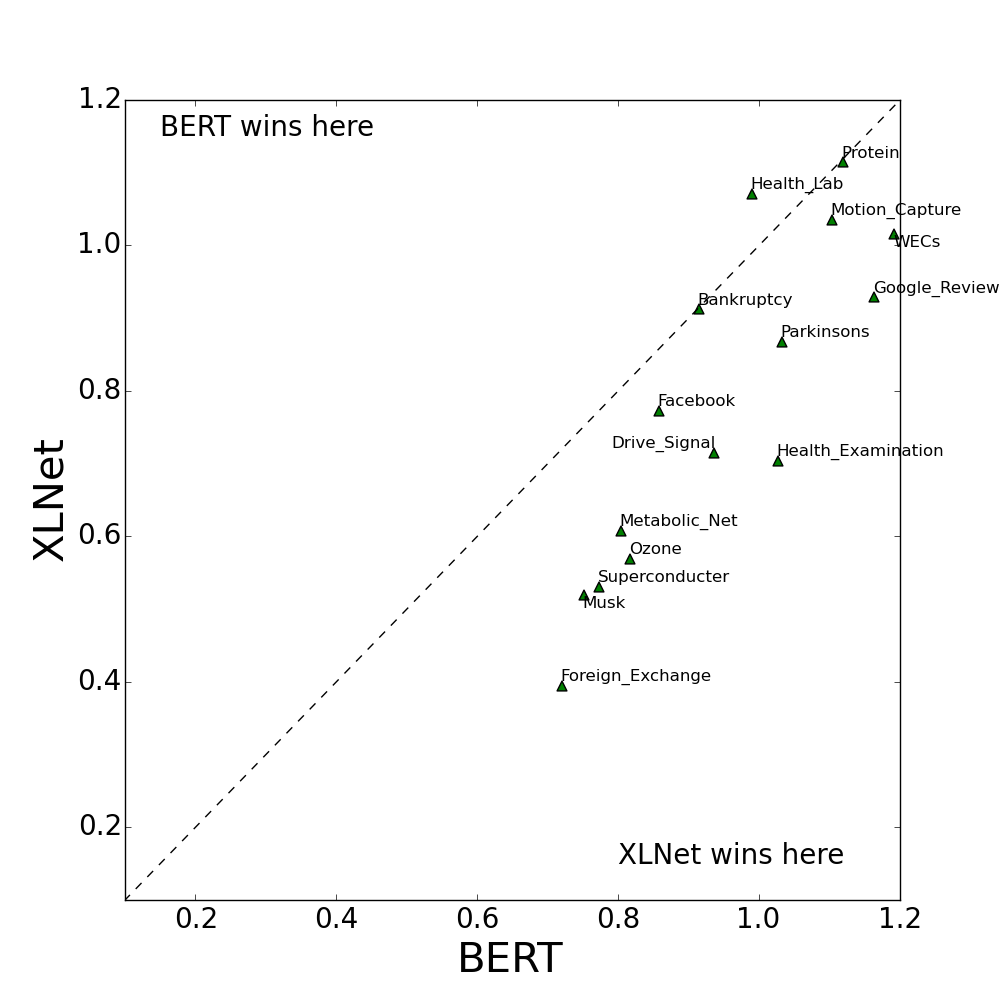}
    \end{subfigure}%
    \caption{Comparison of WNRMSE for Chordalysis \textit{versus} VAEAC, VAEAC \textit{versus} XLNet and BERT \textit{versus} XLNet on the 15 \textbf{continuous} datasets.}\label{fig:wnrmse_scatter_plot}
\end{figure}

Despite all the similarity, there is one interesting new finding we have observed from the experiments on the continuous data; Chordalysis that deals with the continuous attributes by discretizing them and then infers their joint distribution as a graphical model turns out to be a robust model under the highest masking rate. This is evidenced by the increasing mean rank of Chordalysis from 3.87 at the 20\% masking rate to 3.0 at the 80\% while surpassing VAEAC, BERT and MADE along the way. This indicates that Chordalysis tends to generalize its performance better towards highly masked data compared to the above models. The reason behind might be the conservative behaviour of the log-linear analysis performed by Chordalysis on the chordal graphical models \citep{petitjean2013scaling,PetitjeanEtAl14a}. 

Even under the 20\% masking rate, we can still see that Chordalysis is close to the second-best model BERT in the mean rank with the absolute difference being only 0.53. Furthermore, both Chordalysis and XLnet work on discretized data and manage to achieve desirable performance under high masking rates. These observations inform us that discrete models can still perform well on continuous data given proper modelling assumption and regularization; Chordalysis, as a graphical model, has the potential of working as well as some of the best-performing deep learning models under proper modification.

For the other models that mainly focus on minimizing a MSE loss function, MADE works relatively well across the different masking rates with the absolute mean rank difference being only 0.6 between the 20\% and 60\% rates. Again, this can be attributed to the permutation-based autoregressive modelling adopted by MADE. It is also somewhat surprising to see that GAIN and MIDAS perform worse than the Median method frequently across the different masking rates. This suggests that they might not be suitable for target-agnostic prediction tasks on continuous data.

Figure \ref{fig:wnrmse_scatter_plot} shows the pairwise WNRMSE plots that compare the different best-performing models over the continuous data. For Chordalysis versus VAEAC, we can see that Chordalysis starts to acquire more wins than VAEAC when the masking rate for the test data increases from 20\% to 80\%. Even for the loss cases, the performance margins between the two on most of them becomes smaller with higher masking rates. This aligns with our observation on the CD diagrams in Figure \ref{fig:cd_digrams_cont} that Chordalysis overtakes VAEAC in the mean rank at the largest masking rate. For VAEAC versus XLNet, we have a similar finding as that obtained from the CD diagrams; XLNet surpasses VAEAC with increasing masking rates, in this case, on the number of wins, which is also the case for BERT versus XLNet.

\subsection{Results on discretized Data}
\begin{figure}[!htbp]
\centering  
\includegraphics[width=6in]{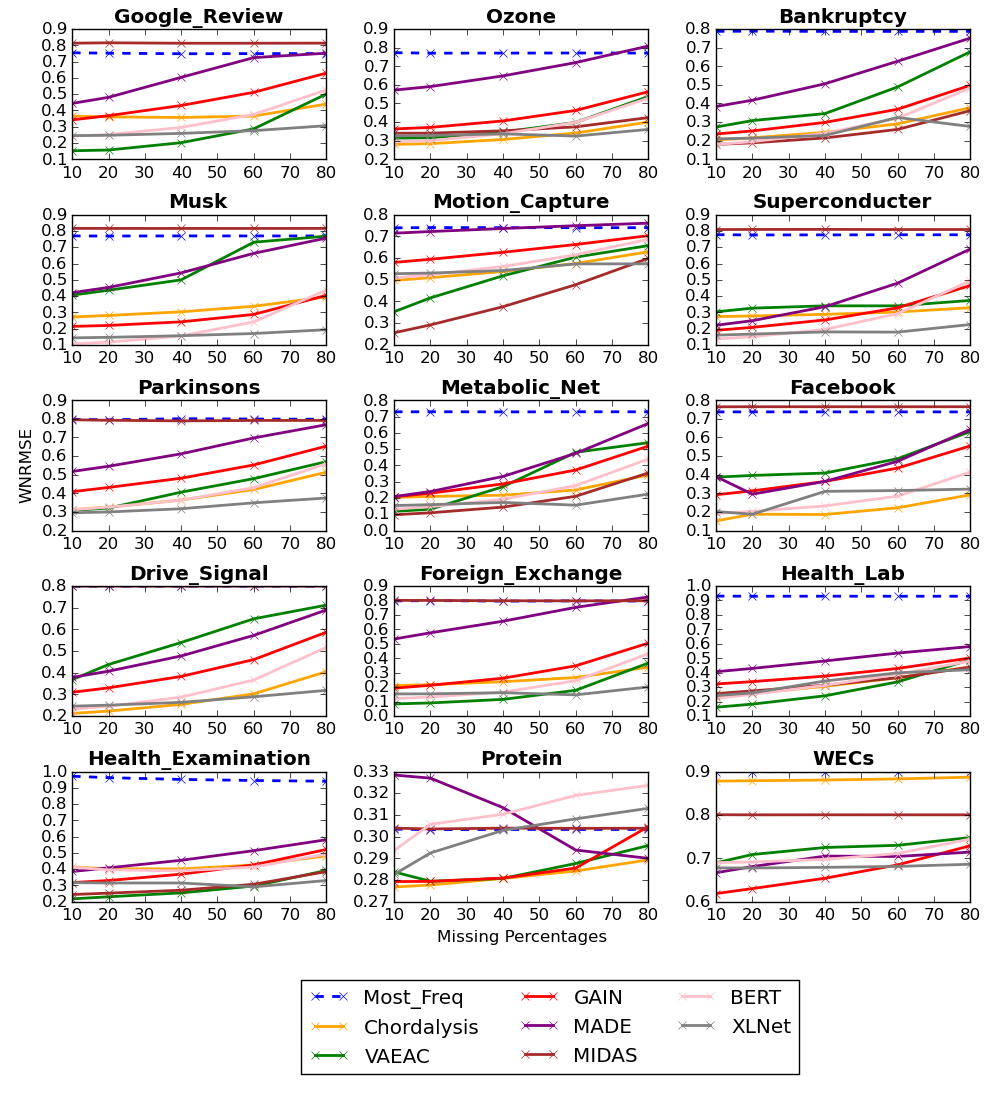}
\caption{The WAPMC results of each model on the different discretized datasets under the different masking rates for their test datasets.}\label{fig:wapmc_dis_results}
\end{figure}

Given the results of each model on the categorical and continuous data, a further investigation is how each model might perform differently on the disretised data. To do so, we discretize all the continuous datasets with 5-bin equal-frequency discretization. Figure \ref{fig:wapmc_dis_results} presents the performance of each model on these discretized datasets. Comparing with Figure \ref{fig:wnrmse_resuts}, we can see that the best (worst) model on a continuous dataset is not necessarily the best (worst) one on the corresponding discretized dataset. With that being said, the overall top-4 models have not changed which are VAEAC, Chordalysis, BERT and XLNet. The only exception is on the discretized WECs dataset where Chordalysis has performed notably worse than the other three but better than BERT and XLNet on the original WECs. This indicates that on this discretized dataset, Chordalysis tends to make (bin number) predictions that are close to (e.g. directly smaller or greater than) the ground-truth bin numbers. In addition, GAIN and MIDAS have performed relatively better on the discretized data than on the continuous data, while MADE has performed worse.

\begin{figure}[!htbp]
    \begin{subfigure}[t]{0.95\textwidth}
        \centering
        \includegraphics[height=2.2in]{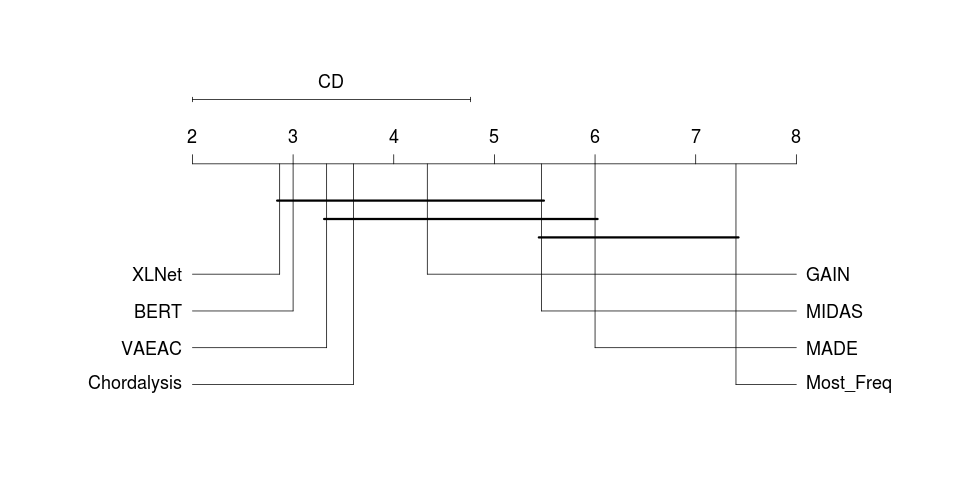}
        \vspace*{-25pt}\caption{The CD diagram for 20\% masking rate}
    \end{subfigure}%
    \\
            \begin{subfigure}[t]{0.95\textwidth}
        \centering
        \includegraphics[height=2.2in]{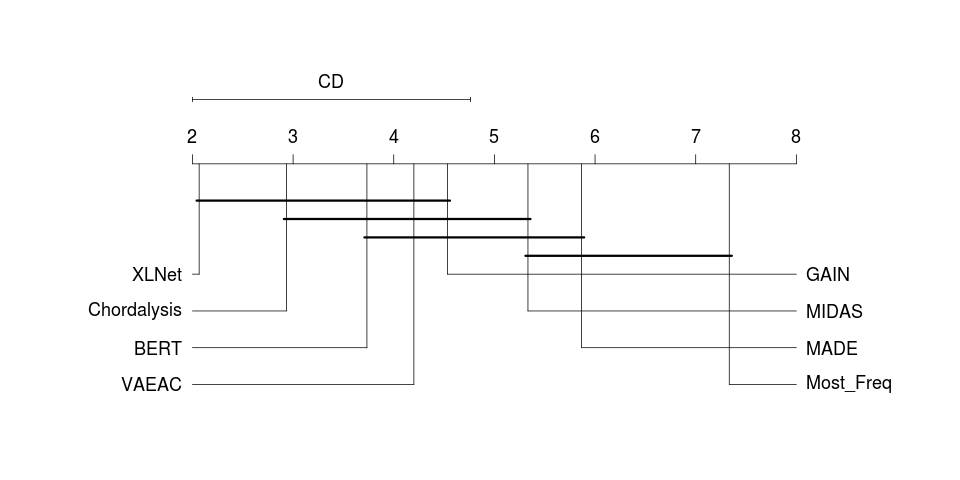}
        \vspace*{-25pt}\caption{The CD diagram for 60\% masking rate}
    \end{subfigure}%
    \\
            \begin{subfigure}[t]{0.95\textwidth}
        \centering
        \includegraphics[height=2.2in]{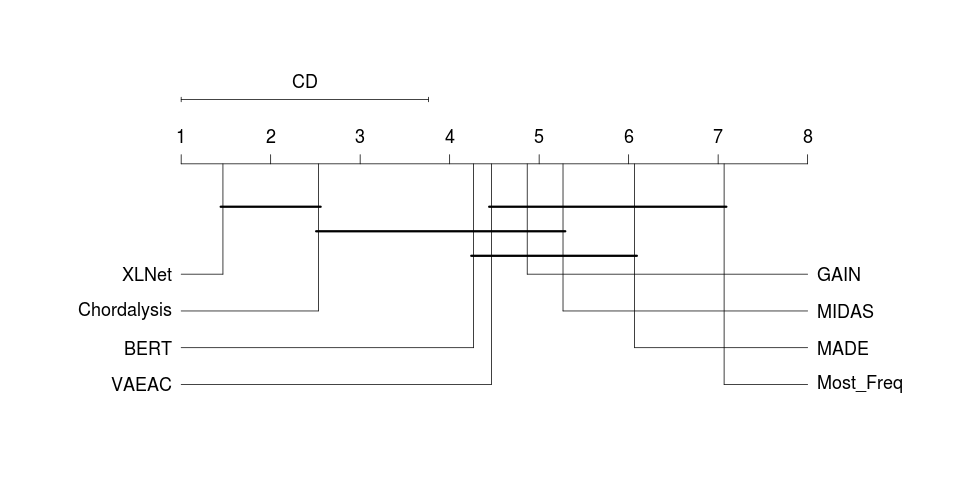}
        \vspace*{-25pt}\caption{The CD diagram for 80\% masking rate}
    \end{subfigure}%
    \caption{The critical difference (CD) diagrams show the average ranks of each model under the different masking rates across the different \textbf{discretized} datasets.}\label{fig:cd_digrams_dis}
\end{figure}

The above observations can also be found from the CD diagrams in Figure \ref{fig:cd_digrams_dis}. Especially, the mean ranks of Chordalysis and XLNet have respectively increased from 3.6 and 2.87 at the 20\% masking rate to 2.53 and 1.47 at the 80\%. In comparison, the mean ranks of VAEAC and BERT have respectively declined from 3.33 and 3.0 at the 20\% to 4.47 and 4.27 at the 80\%. These findings align with our previous conclusion that VAEAC and BERT are less robust than XLNet and Chordalysis when the masking rate for the test data increases. 

Note that for VAEAC, its mean ranks have been relatively lower on the discretized data than on the other types of data under the same masking rates. This might be because the latent representations drawn from the  univariate Gaussian modelled by the encoder of VAEAC are too restrictive to capture the ordinal nature of the attributes as effectively as XLNet and Chordalysis. Also from Figure \ref{fig:cd_digrams_dis}, GAIN and MADE have swapped their mean rank positions compared to their positions on the other types of data even though the difference is not statistically significant.

\begin{figure}[!htbp]
\centering
    \begin{subfigure}[t]{0.4\textwidth}
        \centering
        \includegraphics[height=2.5in]{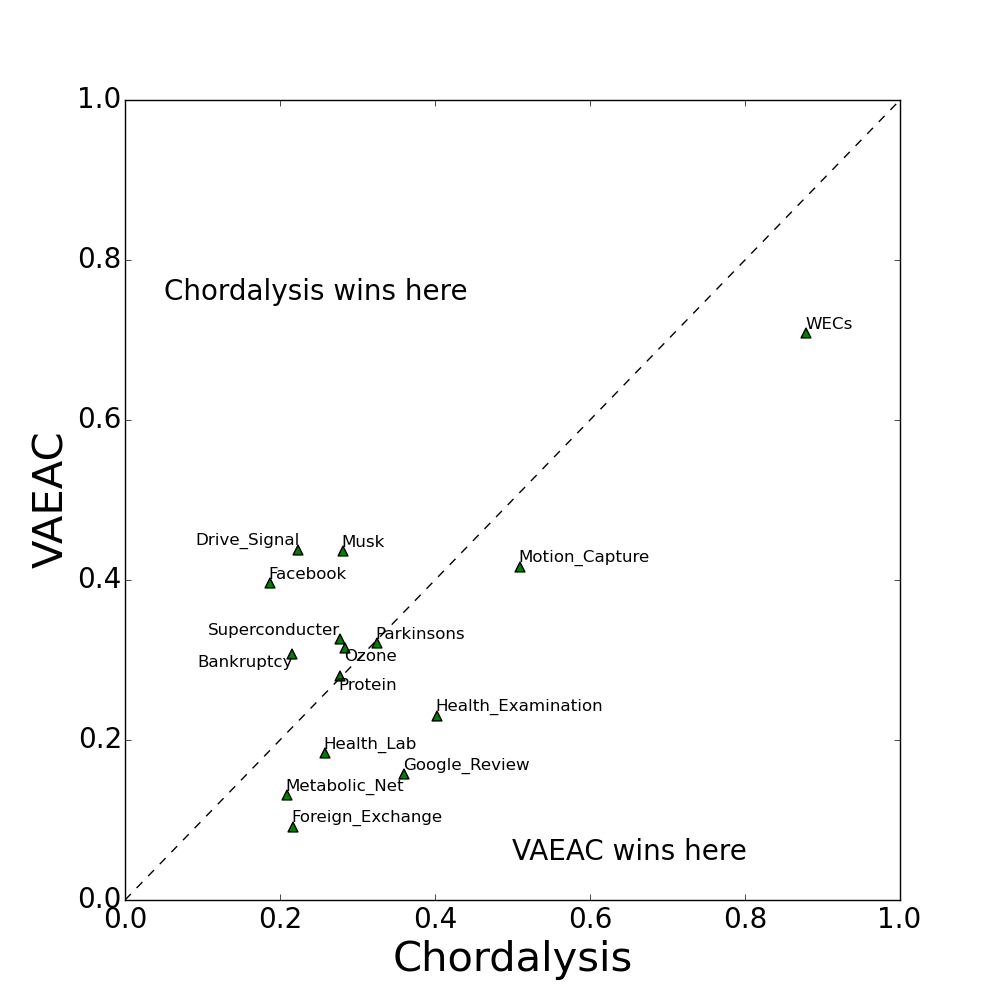}
    \end{subfigure}%
        \begin{subfigure}[t]{0.4\textwidth}
        \centering
        \includegraphics[height=2.5in]{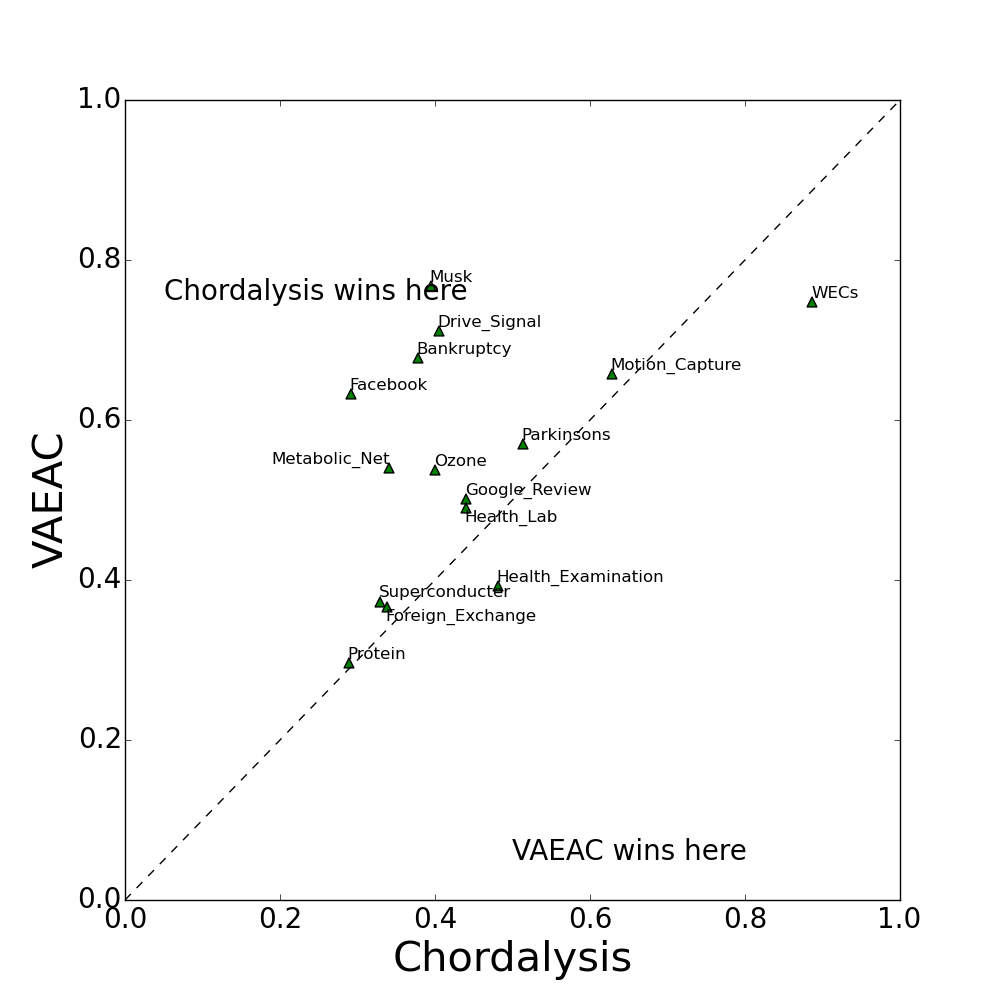}
    \end{subfigure}%
    \\
        \begin{subfigure}[t]{0.4\textwidth}
        \centering
        \includegraphics[height=2.5in]{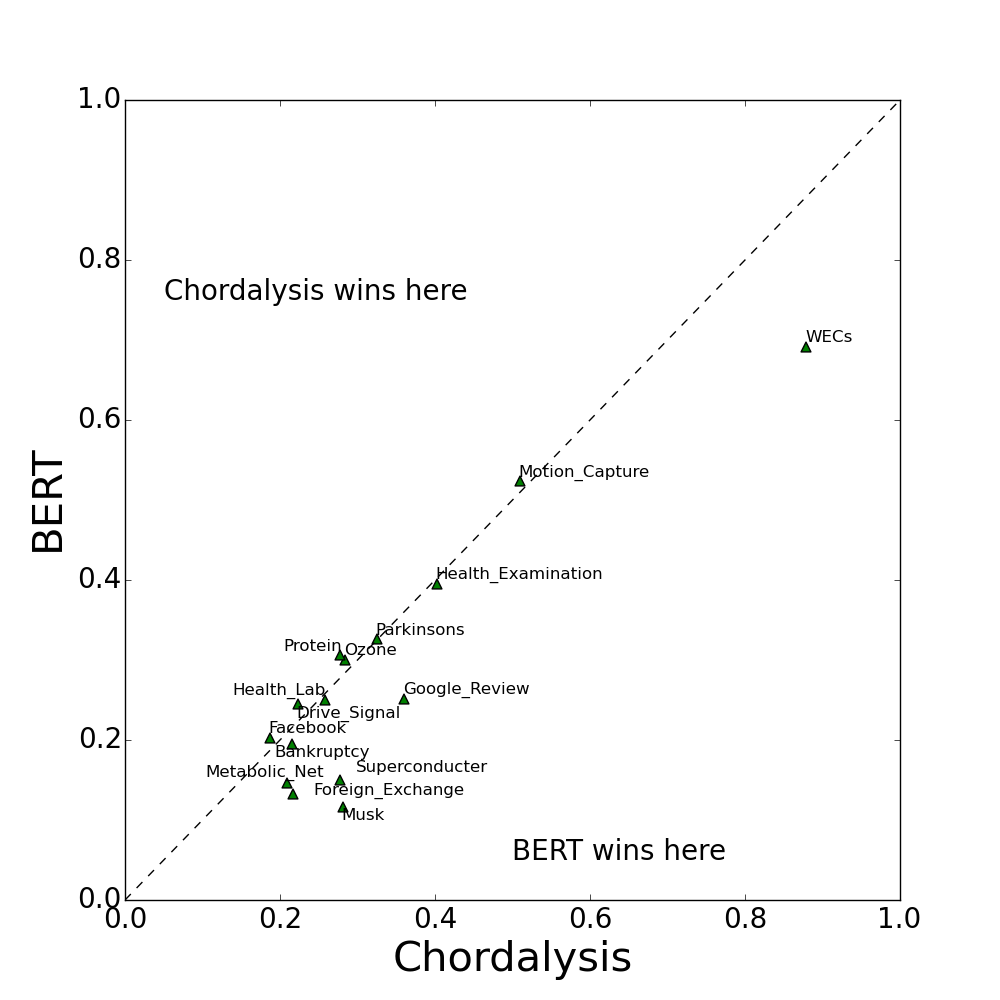}
    \end{subfigure}%
        \begin{subfigure}[t]{0.4\textwidth}
        \centering
        \includegraphics[height=2.5in]{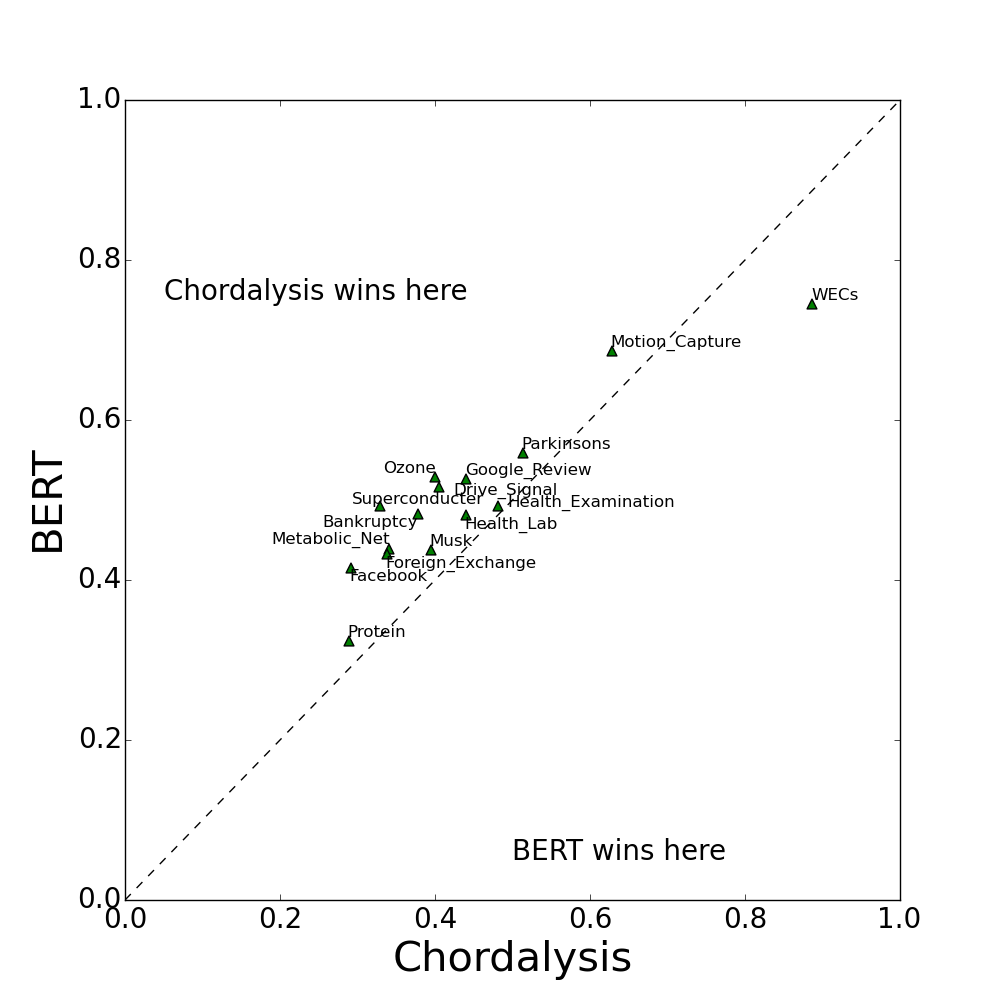}
    \end{subfigure}%
    \\
        \begin{subfigure}[t]{0.4\textwidth}
        \centering
        \includegraphics[height=2.5in]{images/BERT_XLNet_perform_scatter_20_num.png}
    \end{subfigure}%
        \begin{subfigure}[t]{0.4\textwidth}
        \centering
        \includegraphics[height=2.5in]{images/BERT_XLNet_perform_scatter_80_num.png}
    \end{subfigure}%
    \caption{Comparison of WAPMC for VAEAC \textit{versus} Chordalysis, BERT \textit{versus} Chordalysis and XLNet \textit{versus} BERT on the 15 \textbf{discretized} datasets.}\label{fig:wapmc_dis_scatter_plot}
\end{figure}

Figure \ref{fig:wapmc_dis_scatter_plot} displays the pairwise WAPMC comparisons of the different best-performing models across the discretized datases. For Chordalysis, it started by having more losses than VAEAC (i.e. 6 wins, 8 losses and 1 tie) and BERT (i.e. 5 wins, 9 losses, and 2 ties) under the 20\% masking rate. It ends up dominating in the number of wins compared to the other two (i.e. 13 wins and 2 losses against VAEAC and 14 wins and 1 loss against BERT) under the 80\% rate. For BERT versus XLNet, at the 20\% masking rate, the former is slightly better with 6 wins, 3 losses and 6 ties, while at the 80\%, the latter surpasses it by significant margins with 14 wins and 1 loss. This corresponds to the finding from Figure \ref{fig:cd_digrams_dis} that the performance of XLNet is statistically better than that of BERT at the 80\% masking rate.

\subsection{Discussion}
Based on the above experiments performed on the three types of data, we have found that overall, VAEAC, Chordalysis, BERT and XLNet are the most competitive models for target-agnostic learning. Among them, VAEAC and BERT tend to work better when the test data is highly or moderately observed, while Chordalysis and XLNet are more robust towards highly unobserved test data. Here, we summarize and posit the following conjectures that can be responsible for the above observations:
\begin{itemize}
    \item VAEAC has not scaled its KL-divergence term to properly regularize its latent representations from the encoder to deal with the scenario where the training data is highly observed but the testing data is highly masked. Scaling the KL-divergence term has been shown to improve the generalization of variational autoencoders for various applications \citep{Higgins2017betaVAELB,liangVAEforCF2018}. In addition, for discretized data, the univariate Gaussian latent representations generated by VAEAC might not be able to fully capture the orderings in the values of each attribute.
    \item BERT imposes an independence modelling assumption on the prediction of the masked tokens. This might have limited the generalization ability of its target-agnostic learning in predicting for a large set of target attributes based on a small set of observed attributes.
    \item In comparison, XLNet has leveraged permutation-based autoregressive modelling that tends to better exploit the dependency between the observed and masked attributes. This, along with the stacked transformer-XL architecture, is likely to be the reason for XLNet to perform robustly on the highly masked (test) data.
    \item Chordalysis has performed not only the most effectively on the categorical data but also robustly on the continuous data with simple equal-frequency discretization and bin-median prediction. This shows that discrete graphical models based on scalable chordal graph decomposition can perform equally well as the deep learning models on target-agnostic learning for any type of attributes.
\end{itemize}

%

\subsection{Time Complexity Analysis}
\begin{figure}[!htbp]
\centering  
\includegraphics[width=6in]{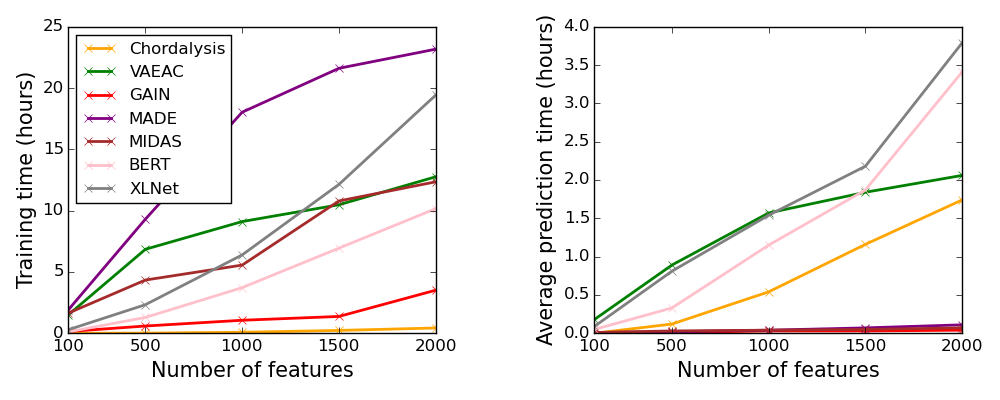}
\caption{The training and average prediction time (under the 80\% test masking rate) of different models in hours against the number of features on the NYT-2000 dataset.}\label{fig:time_nyt}
\end{figure}

Now that we have studied the performance of each model across the different types of attributes under a wide range of masking rates for the test data, we would like to further investigate the computational complexity of each model in different scenarios. For the following experiments, we run each deep learning model on the same machine, the configuration of which has been specified in Section \ref{sec:experiment_settings}, with the same batch size which is 64 for both training and testing. We start by investigating how the different models scale with respect to the number of attributes. Figure \ref{fig:time_nyt} shows both the training time under the 20\% masking rate for the training data, and the average prediction time over the 10 imputations under the 80\% masking rate for the test data. For this particular experiment, we use the NYT-2000 dataset because 1) it is a large-scale data with 185,771 data instances and 2,000 attributes and 2) its attributes are ordered by the number of occurrences of each word, which allows us to study the model performance with an increasing number of attributes. 

With respect to the training time, we can clearly see that Chordalysis is the fastest to finish the training. When the full set of attributes (i.e. 2,000 attributes) is used, the training of Chordalysis still finishes within one hour. The second-best model is GAIN whose training time has the same order of magnitude as Chordalysis and finishes its training within 4 hours with the full set of attributes. For the other models, their training time is at least an order of magnitude slower than that of Chordalysis. The two slowest models in this case are MADE and XLNet which take nearly a day to train. This is due to not only their model complexity but also the permutation of the inputs they adopt for their training. When only 100 attributes are used, VAEAC is notably slower than Chordalysis (i.e. by 1.48 hours), BERT (i.e. by 1.37 hours) and XLNet (i.e. by 1.20 hours) while its training time experienced an almost logarithmic growth from 100 to 2,000 attributes.

As for the testing/prediction time, when only 100 attributes are used, there is no significant difference in each model. The difference becomes notable as the number of attributes increases to 500. With the full set of attributes, GAIN is the average fastest model to finish one round of imputation for the test data in 0.04 hour. MIDAS and MADE are the second and third models to finish the imputation with 0.07 and 0.11 hour respectively even though their training is massively slower. It is interesting to see that they are slow to train but fast to perform the prediction. We conjecture that this is because their back-propagation during the training is slowed down by the dropout (for MIDAS) and the weight-masking (for MADE) operations. For the testing, there is no back-prograpation and only a single forward pass of their networks is needed to obtain each imputation for the test data.   

For Chordalysis, given the limitation of our computational resources, its prediction can maximally be parallelized over 10 chunks of data subsets. In this case, it spent on average 1.74 hours to finish each imputation with the full set of attributes. We believe that with a larger number of processes or threads, the prediction of Chordalysis will be further accelerated. The biggest bottleneck of its computation comes from the hardware; its computation is based on CPUs and is hard to be migrated to GPUs. In addition, the prediction of VAEAC, BERT and XLNet is notably slower than the top-3 models in this case, most likely due to their higher model complexity. It is also interesting to see that XLNet is significantly slower than MADE despite that they both randomly permute their inputs for the multiple imputation. We conjecture that the difference is caused by XLNet's complex two-stream self-attention mechanism and the memory mechanism over the stacked transformer-XL architecture.

\begin{figure*}[!htbp]
\centering  
\includegraphics[width=6in]{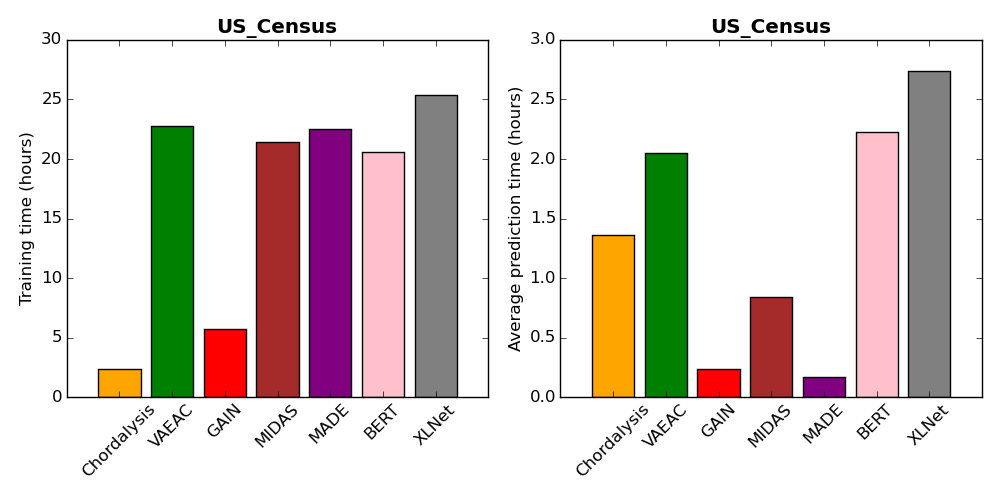}
\caption{The training and average prediction time (under the 80\% test masking rate) of different models on the largest-quantity categorical dataset ``US\_Census".}\label{fig:time_us_census}
\end{figure*}

\begin{figure*}[!htbp]
\centering  
\includegraphics[width=6in]{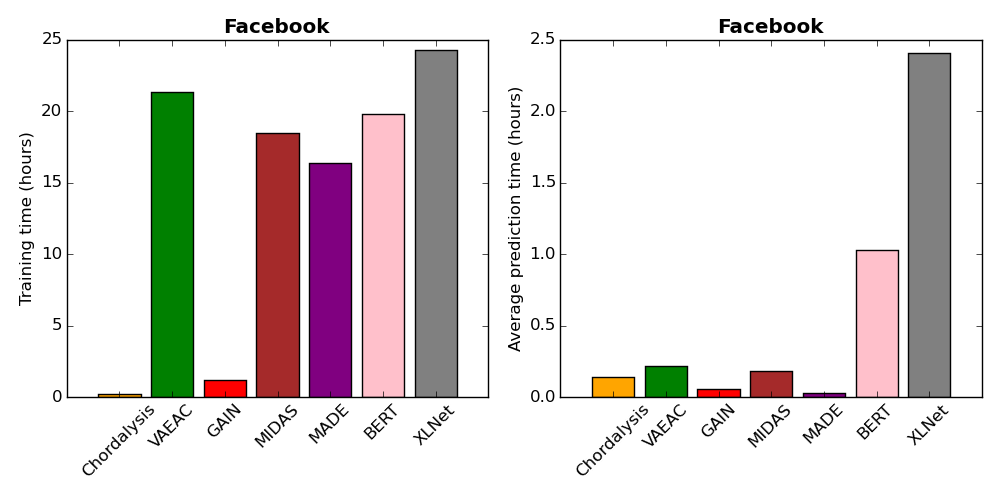}
\caption{The training and average prediction time (under the 80\% test masking rate) of different models on the largest-quantity continuous dataset ``Facebook".}\label{fig:time_facebook_census}
\end{figure*}

Next, we proceed to studying the training and average prediction time (under the 80\% masking rate) of each model on both the categorical/discretized data and the continuous data. We only report the results on the largest categorical and continuous datasets, which are respectively the US\_Census and the Facebook, as the results on the other smaller datasets share similar patterns. Figures \ref{fig:time_us_census} and \ref{fig:time_facebook_census} show the training and average prediction time of each model, respectively, over the US\_Census and the Facebook data. We can see that Chordalysis has spent the lowest training time on both datasets. This shows that the training of Chordalysis is the least affected by either the quantity of data or the number of attributes. The same observation has been found on GAIN, which is second to Chordalysis in the training time for both scenarios. In comparison, the other models are significantly slower in training and their amounts of training time are also comparable to each other on both datasets. 

As for the testing time, MADE and GAIN are the top-2 models over both datasets and there is almost no difference in their performance. Therefore, these two models are the least affected by either the quantity of data or the number of attributes for target-agnostic prediction. We can also see that the prediction time of VAEAC, Chordalysis, BERT and XLNet is comparable on the categorical data, while on the continuous data, the former two are notably faster than the latter ones. Furthermore, the prediction time of VAEAC and Chordalysis on the continuous data is comparable to that of GAIN, MADE and MIDAS. For VAEAC, this is mostly because its output dimension has reduced significantly compared to that for the categorical and discretized data, which results in the reduction in the model and computational complexity. For Chordalysis, BERT and XLNet, they all work by discretizing the data and making prediction based on the bin medians. The difference in their prediction time, in this case, is solely determined by their model complexity for which Chordalysis is the simplest and XLNet is the most complex.

\section{Conclusion and Future Work}
The machine learning community has not previously considered the task of predicting any attribute from any available information in any scenario. As a result, there lacks a systematic body of techniques for understanding and addressing this specific type of problem. In this paper, we have formally introduced \textit{target-agnostic} learning, which is the key solution to such problems. It provides a general learning paradigm that encompasses  \textit{generative}, \textit{discriminative} and \textit{self-supervised} learning. 

More specifically, target-agnostic learning can be addressed by generative systems that learn a model of the full joint distribution that captures the interdependence among all the attributes. In this case, the key to drawing predictions for any arbitrary set of target attributes is efficient marginalization. Graphical models \citep{koller2009probabilistic} achieve this with their efficient inference algorithms, which enable marginalization of the joint distribution over any set of attributes. Chordalysis is a state-of-the-art graphical model learning framework for large data. It leverages the decomposibility of chordal graphs for scaling up the learning of graphical models with thousands of attributes in seconds and meanwhile enables efficient and exact marginalization over any attributes. In this paper, we have focused our investigation on Chordalysis as an effective exemplar of the generative approach for target-agnostic learning.

As for discriminative learning, it may not seem to be compatible with target-agnostic learning as its traditional approaches focus on predicting (the conditional probabilities of) single target attributes. With that being said, we show that target-agnostic learning can still be discriminative via the pseudo-likelihood theory \citep{besag1975statistical}. More specifically, we prove that the pseudo-likelihood, as a product of conditional probabilities for each attribute given all the others, approaches the joint distribution in general as the number of data instances approaches infinity. In other words, discriminative learning can implicitly fit a joint distribution model, which can be viewed as a form of target-agnostic learning. More importantly, the pseudo-likelihood theory elegantly combines the generative and discriminative learning, and provides the theoretical foundation for  target-agnostic learning.

Self-supervised learning can be viewed as a modern redevelopment of pseudo-likelihood theory with neural networks. It aims to learn joint models as certain types of neural networks to generate contextual latent representations for data instances. One way to perform the learning is to maximize the pseudo-likelihood under a target-agnostic prediction task (e.g., a special case in NLP is the random masked token prediction for sentences). The self-supervised learning models that have adopted this particular way of learning, or more generally, by randomly modifying the input attributes for reconstruction, are BERT and XLNet. In this paper, we have adapted these two models from their original NLP domains to successfully dealing with tabular data. Furthermore, for self-supervised learning models that have not adopted target-agnostic prediction for their training, including autoencoders and generative adversarial networks, we have also adapted their most popular methods to such a way of training. To summarize our above contributions, we have:

\begin{itemize}
    \item established relationship of generative learning, discriminative learning and  self-supervised learning to target-agnostic learning via the pseudo-likelihood theory;
    \item specified different ways of building target-agnostic learning algorithms, which include graphical model learning based on chordal decomposition, and pseudo-likelihood maximization under target-agnostic prediction and random modification of input attributes;
    \item adapted BERT and XLNet from their original NLP domains to handling tabular data.
\end{itemize}

For the experiments on target-agnostic learning, we propose two evaluation metrics respectively for the categorical/discretized data and the continuous data, which are the WAPMC and WNRMSE. Both of them weigh each attribute (as the targets) by the different percentages of their masked values across the test instances considering the existence of missing values, and take these weights into account for averaging the errors on each attribute. In addition, WNRMSE also considers the standard deviation of each continuous attribute for normalizing the errors on them. Our experiments were conducted over 15 categorical datasets and 15 continuous (and discretized) datasets, from which we have obtained the following findings:
\begin{itemize}
    \item VAEAC, a VAE model using the target-agnostic training, Chordalysis, BERT and XLNet are the top-performing models across the different types of data; 
    \item VAEAC and BERT perform relatively well on highly observed test data given that they are trained on highly observed data. Their performance deteriorates considerably on highly masked test data given the same training;
    \item XLNet performs not as desirably as VAEAC and BERT on highly observed categorical and continuous (test) data. However, it is clearly more robust than the two when the test data is highly masked. XLNet has also remained the top model on the discretized data (under various extents of masking of their test data);
    \item Chordalysis is also more robust than VAEAC and BERT towards the highly masked continuous and discretized (test) data, while it remains the best model on the categorical data. This shows that graphical models with scalable learning and effective marginalization can perform as good as the state-of-the-art deep learning models on target-agnostic prediction tasks.
\end{itemize}

We then proceeded to studying the training and prediction time complexity for each model in two scenarios: on a large number of attributes and on a large quantity of data. Correspondingly, we have obtained the following findings:
\begin{itemize}
    \item The training time of Chordalysis is an order of magnitude faster than that of the other models except GAIN on a large-scale dataset with 2,000 attributes; Its prediction is intrinsically sequential and can be further accelerated via multi-processing or multi-threading parallel computation;
    \item The prediction time of VAEAC on a large-quantity continuous data is comparable to that of the fastest models in this regard; This is not the case on a large-sized categorical data;
    \item Overall, VAEAC, BERT and XLNet are notably slower in training compared to the other models for either large-quantity or large-scale data regardless of the types of the data. BERT and XLNet are also notably slower in target-agnostic prediction over the test data.
\end{itemize}

For future work, we will focus on perfecting the current best-performing target-agnostic learning models by addressing their underperforming scenarios. For VAEAC, we conjecture that its unscaled KL-divergence has contributed to its low performance on highly masked data. Therefore, we would like to leverage the $\beta$-VAE \citep{Higgins2017betaVAELB} architecture to address this issue. The scaling term $\beta$ will balance the trade-off between the pseudo likelihood and the KL-divergence. We can either make $\beta$ a hyper-parameter or learn it together with the other network parameters by modelling it dependent on the observed data and masks.

For BERT, we conjecture that its independence assumption on the masked token prediction has adversely affected its performance on highly masked data. Despite that XLNet is a better option in this case, we can still leverage pre-training for BERT to improve its performance. We envisage the pre-training to be domain-specific. For example, we can use the tabular attributes of some large medical health databases from an open-source project (e.g. the MIMIC project) to pre-train a BERT model based on our target-agnostic adaptation for the tabular data. After the pre-training, the target-agnostic fine-tuning or prediction of the model can be performed on any following health databases from the same project. 

To accelerate the training and/or prediction of BERT, we can leverage the DistilBERT \citep{Sanh2019DistilBERTAD}, a lighter and faster version of BERT with slightly lower predictive performance. It exploits the knowledge distillation technique \citep{HintonKnowDistill2015} that uses a large pre-trained BERT model (i.e. one pre-trained on a large-scale dataset) to ``teach" a smaller BERT with a reduced number of hidden layers (using a much smaller training subset from the same dataset) to perform comparably and in a similar way. 

For Chordalysis, there are two interesting research directions for us to investigate. The first one is to further improve the learning of the graphical model structures. This involves developing new scoring methods for graphical models that incorporate constraints specific to the trade-off between goodness of fit on the data and model complexity. The other direction will focus on how best to estimate model parameters that will maximize the prediction accuracy on any arbitrary set of target attributes given a particular model structure. A recent advance in leveraging hierarchical Dirichlet processes for smoothing and regularizing the conditional probability tables of Bayesian network classifiers has enabled the classifiers to achieve predictive performance on a single target attribute that is competitive with the state-of-the-art discriminative learning algorithms \citep{petitjean2018accurate}. We will seek to generalize this capability from being target-specific to being target-agnostic.

\clearpage

\subsection*{Acknowledgements}
This work was supported by the Australian Research Council under awards
DE170100037.
Wray Buntine and Geoff Webb are also  sponsored by DARPA under agreement number FA8750-19-2-0501. The U.S. Government is authorized to reproduce and distribute reprints for Governmental purposes notwithstanding any copyright notation thereon. The authors also would like to thank Dr. Mark James Carman for providing valuable suggestions and feedback to this work.

\appendix
\section*{Appendix}

\begin{figure}[!htbp]
    \begin{subfigure}[t]{0.95\textwidth}
        \centering
        \includegraphics[height=2.2in]{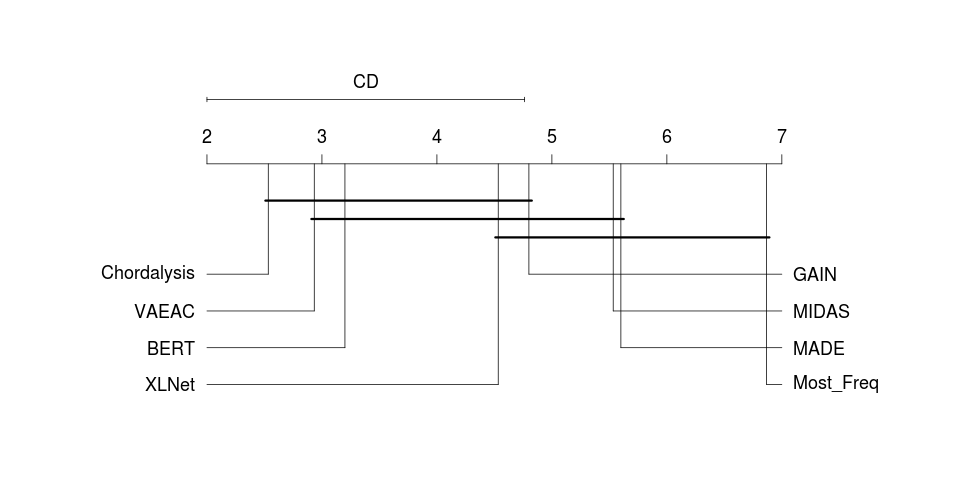}
        \vspace*{-25pt}\caption{The CD diagram for 10\% masking rate}
    \end{subfigure}%
    \\
            \begin{subfigure}[t]{0.95\textwidth}
        \centering
        \includegraphics[height=2.2in]{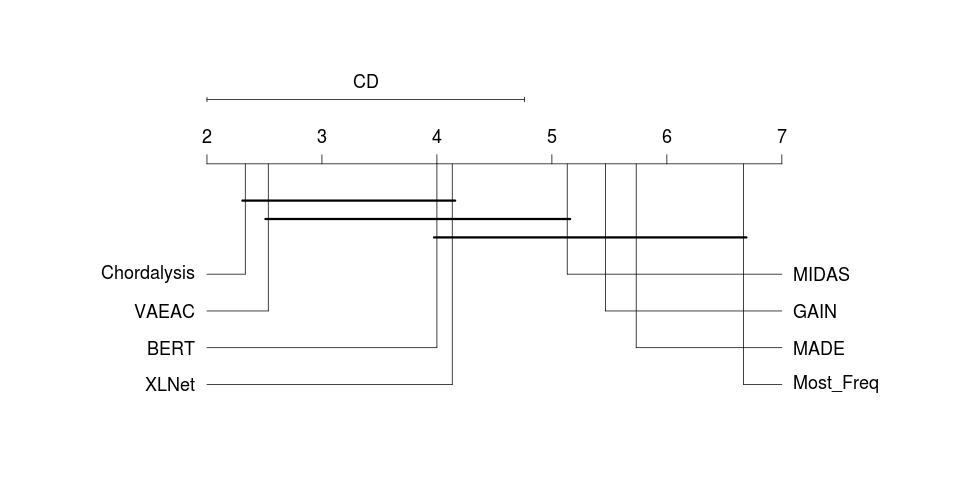}
        \vspace*{-25pt}\caption{The CD diagram for 40\% masking rate}
    \end{subfigure}%
    \caption{The critical difference (CD) diagrams show the average ranks of each model under the 10\% and 40\% masking rates across the 15 \textbf{categorical} datasets.}\label{fig:cd_digrams_cat_append}
\end{figure}

\begin{figure}[!htbp]
    \begin{subfigure}[t]{0.95\textwidth}
        \centering
        \includegraphics[height=2.2in]{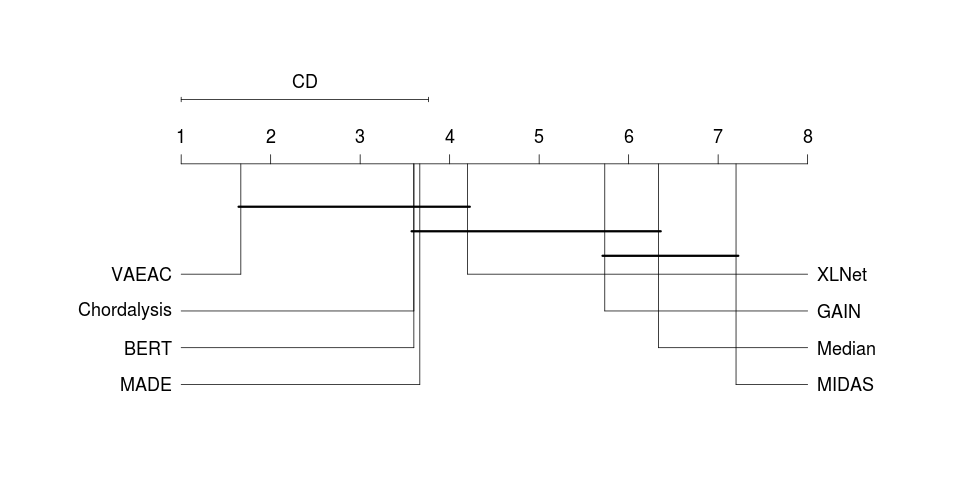}
        \vspace*{-25pt}\caption{The CD diagram for 10\% masking rate}
    \end{subfigure}%
    \\
            \begin{subfigure}[t]{0.95\textwidth}
        \centering
        \includegraphics[height=2.2in]{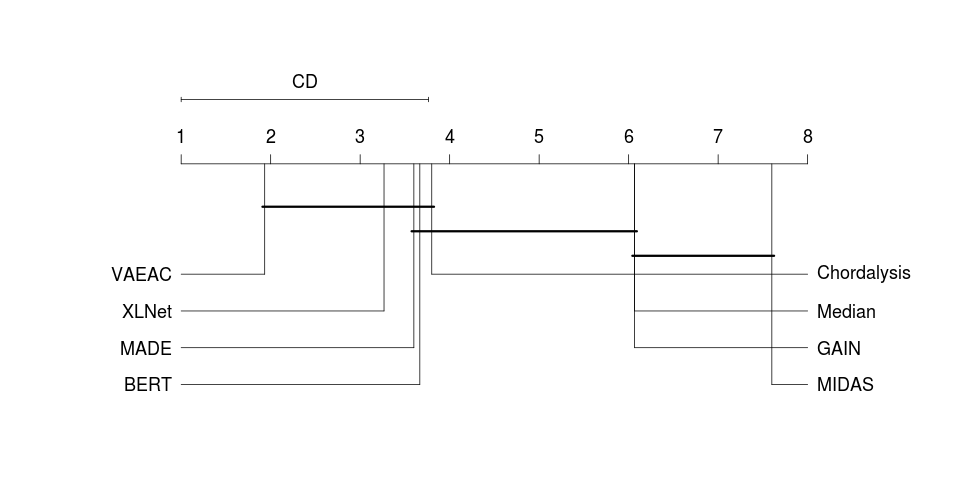}
        \vspace*{-25pt}\caption{The CD diagram for 40\% masking rate}
    \end{subfigure}%
    \caption{The critical difference (CD) diagrams show the average ranks of each model under the 10\% and 40\% masking rates across the 15 \textbf{continuous} datasets.}\label{fig:cd_digrams_cont_append}
\end{figure}

\begin{figure}[!htbp]
    \begin{subfigure}[t]{0.95\textwidth}
        \centering
        \includegraphics[height=2.2in]{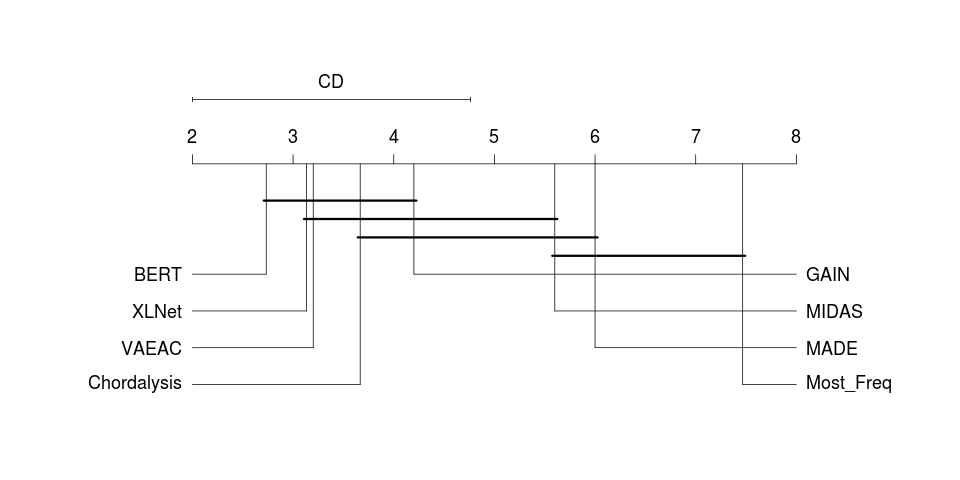}
        \vspace*{-25pt}\caption{The CD diagram for 10\% masking rate}
    \end{subfigure}%
    \\
            \begin{subfigure}[t]{0.95\textwidth}
        \centering
        \includegraphics[height=2.2in]{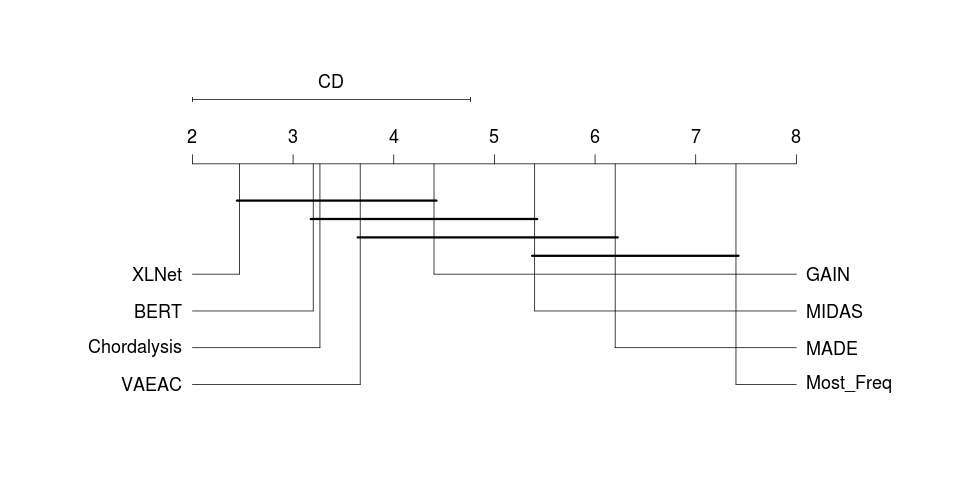}
        \vspace*{-25pt}\caption{The CD diagram for 40\% masking rate}
    \end{subfigure}%
    \caption{The critical difference (CD) diagrams show the average ranks of each model under the 10\% and 40\% masking rates across the 15 \textbf{discretized} datasets.}\label{fig:cd_digrams_dis_append}
\end{figure}

See Figures \ref{fig:cd_digrams_cat_append}, \ref{fig:cd_digrams_cont_append} and \ref{fig:cd_digrams_dis_append}.


\clearpage

\bibliography{bibliography}

\end{document}